\documentclass[a4paper]{article}

\usepackage{amsmath,amssymb,amsfonts}
\usepackage{bm}
\usepackage{graphicx}
\usepackage{booktabs}
\usepackage{algorithm}
\usepackage{algorithmic}
\usepackage{siunitx}
\usepackage{url}
\usepackage{authblk}

\begin{document}

\title{Gauss--Hermite Quadrature for Gaussian-Mixture Entropy\\
with an Action-Space Hermite Surrogate}

\author[1,2,3]{Jae Wan Shim}

\affil[1]{Extreme Materials Research Center, Korea Institute of Science and Technology, 5 Hwarang-ro 14-gil, Seongbuk, Seoul, 02792,  Republic of Korea}

\affil[2]{Climate and Environmental Research Institute, Korea Institute of Science and Technology, 5 Hwarang-ro 14-gil, Seongbuk, Seoul, 02792,  Republic of Korea}

\affil[3]{Division of AI-Robotics, KIST Campus, University of Science and Technology, 5 Hwarang-ro 14-gil, Seongbuk, Seoul, 02792,  Republic of Korea}

\maketitle

\begin{abstract}
Gaussian distributions are used to model uncertainty in signals and states,
and Gaussian mixtures are often used when the underlying distribution is
multimodal. Unlike a single Gaussian, a Gaussian mixture generally has no
closed-form expression for differential entropy and therefore requires
numerical approximation. We evaluate Gaussian mixture differential entropy
using direct componentwise Gauss--Hermite quadrature. The quadrature order controls 
the numerical resolution of the approximation.
The method is evaluated on one- and two-dimensional Gaussian mixture
benchmarks against Taylor approximations, analytic entropy bounds, and
numerical integration references.

For repeated optimization over continuous actions, we construct a
Hermite polynomial surrogate in action space. In a closed-loop radar sensor-management benchmark, 
its second-order form achieves substantially
lower surrogate error and optimizer regret than a second-order Taylor
surrogate based on local derivatives at the nominal action, while both methods
use nine direct objective evaluations per replanning step. 
The Hermite surrogate also improves detection, inside-gate, and
beam-pointing metrics in the tested benchmark.
\end{abstract}


\section{Introduction}
Gaussian mixture models (GMMs) provide a flexible representation of
non-Gaussian and multimodal probability distributions. They are widely used in
target tracking, mapping, and Bayesian state estimation, where multimodality
may arise from uncertain data association, nonlinear measurement
relationships, or multiple competing hypotheses
\cite{Salmond1989IEEColloquium,HanebeckBriechleRauh2003}. Quantifying the
uncertainty represented by such distributions is important for estimation and
decision making. Entropy, cross-entropy, mutual information, and
Kullback--Leibler divergence are commonly used for this purpose in
information-driven sensing, planning, control, mixture comparison, and mixture
reduction
\cite{Shannon1948,CoverThomas1991,ManyikaDurrantWhyte1992,ViolaWells1997,
KullbackLeibler1951,HersheyOlsen2007,GoldbergerGordonGreenspan2003,Runnalls2007}.

Information-driven sensor management uses predicted estimation
performance to select sensing actions and allocate sensing resources.
Recent work has considered such decision making in target search and
tracking, including multi-target sensor management and nonmyopic sensor
control~\cite{BostromRostEtAl2021,HernandezEtAl2024,JonesEtAl2024}.
These problems motivate computationally efficient evaluation and
optimization of uncertainty-based objectives when the underlying belief
is represented by a Gaussian mixture.

Unlike the single-Gaussian case, Gaussian-mixture differential entropy
generally has no closed-form expression and must be evaluated approximately.
Existing approaches include Taylor expansions about the means of the Gaussian components,
analytic upper and lower bounds, direct
numerical integration, Monte Carlo integration, and polynomial approximations.
Huber \textit{et al.}~\cite{Huber2008MFI} developed a representative
Taylor construction based on expansions about the means of the Gaussian
components. Maximum-entropy arguments provide analytic bounds
\cite{NielsenNock2017}. Dahlke and Pacheco analyzed the
convergence of polynomial entropy approximations and showed that a commonly
used Taylor expansion about the means of the Gaussian components can diverge
under simple conditions;
they also proposed convergent Taylor and shifted-Legendre alternatives
\cite{DahlkePacheco2023}. Analytic bounds and approximations for mixture entropy have also been studied
extensively. Melbourne \textit{et al.} derived bounds on the differential
entropy of mixtures and on its difference from the weighted component
entropies \cite{MelbourneEtAl2022}. Zobay employed an analytic entropy bound
for Gaussian mixture variational approximations \cite{Zobay2014}.
Furuya \textit{et al.} analyzed a closed-form approximation to Gaussian
mixture entropy based on the component entropies and mixture weights, and
related its accuracy to component separation
\cite{FuruyaEtAl2022EntropyError}.
Gauss--Hermite quadrature has also been used to evaluate Gaussian
log-mixture expectations. Saal \textit{et al.}~\cite{SaalHeessVijayakumar2011}
employed componentwise Gauss--Hermite quadrature for such expectations
in variational Gaussian-mixture filtering and further introduced a
component-dependent reformulation of the quadrature to improve low-order
accuracy for mixtures with substantially different component covariances.
Thus, componentwise Gauss--Hermite evaluation of Gaussian log-mixture
expectations is established prior work. A recent alternative approximates
Gaussian-mixture entropy by fitting a polynomial to the entropy integrand
and exploiting analytically tractable Gaussian-mixture power
integrals~\cite{JoudehSkoric2026}.

Building on this established evaluation technique, the present work makes two contributions. First, we use direct componentwise Gauss–Hermite quadrature as a deterministic state-space entropy evaluator: each Gaussian component is standardized, and the resulting evaluator is validated on one- and two-dimensional benchmark families against Taylor approximations, analytic bounds, and numerical integration references. This validation includes a replication analysis of the one-dimensional example of Huber et al.~\cite{Huber2008MFI}, which identifies an inconsistency between the residual definition printed there and the reported minimizing coefficient. Second, for repeated optimization of the entropy-based continuous-action objective, we construct a Hermite polynomial surrogate in action space. In a closed-loop radar sensor-management benchmark, its second-order form is compared with a second-order Taylor surrogate based on local derivatives at the nominal action, with both methods using the same number of direct objective evaluations. The Hermite surrogate uses objective values over the action neighborhood to compute a Gaussian-weighted projection, whereas the Taylor surrogate is determined by local derivatives at the nominal action.

\section{Problem Formulation}
Let $x \in \mathbb{R}^d$ be a continuous random vector with density $g(x)$. The differential entropy is
\begin{equation}
H(g)
\triangleq
-\int_{\mathbb{R}^d}g(x)\log g(x)\,dx
=
-\mathbb{E}_{X\sim g}\!\left[\log g(X)\right].
\label{eq:entropy_def}
\end{equation}

We consider a Gaussian mixture model (GMM)
\begin{equation}
g(x) = \sum_{j=1}^{L} \omega_j\, \mathcal{N}(x;\mu_j,C_j),
\quad \omega_j > 0,\quad \sum_{j=1}^{L}\omega_j = 1,
\label{eq:gmm}
\end{equation}
Zero-weight components, if present in an input representation, are removed
before evaluation. The quantity $\omega_j$ is the $j$th mixture weight and
$\mathcal{N}(x;\mu_j,C_j)$ denotes the $d$-dimensional Gaussian density with
mean vector $\mu_j \in \mathbb{R}^d$ and positive-definite covariance matrix
$C_j \in \mathbb{R}^{d\times d}$.

Direct evaluation of \eqref{eq:entropy_def} is difficult because the
integrand contains
\begin{equation}
\log g(x)
=
\log
\left[
\sum_{j=1}^{L}
\omega_j
\mathcal{N}(x;\mu_j,C_j)
\right],
\label{eq:log_gaussian_mixture}
\end{equation}
that is, the logarithm of a weighted sum of Gaussian component
densities. This expression generally does not admit analytic
integration with respect to the Gaussian-mixture density.

\section{Entropy Decomposition and Gaussian Standardization}

Substituting \eqref{eq:gmm} into \eqref{eq:entropy_def} and
interchanging the finite sum and the integral yields the following
decomposition; the required integrability is established in
Section~\ref{sec:hermite_representation}:
\begin{equation}
\begin{aligned}
H(g)
&=
-\int_{\mathbb{R}^d}
\left[
\sum_{i=1}^{L}
\omega_i
\mathcal{N}(x;\mu_i,C_i)
\right]
\log g(x)\,dx
\\
&=
-\sum_{i=1}^{L}
\omega_i
\int_{\mathbb{R}^d}
\mathcal{N}(x;\mu_i,C_i)
\log g(x)\,dx
\\
&=
-\sum_{i=1}^{L}
\omega_i
\mathbb{E}_{X_i\sim\mathcal{N}(\mu_i,C_i)}
\left[
\log g(X_i)
\right].
\end{aligned}
\label{eq:component_decomp}
\end{equation}
This decomposition is exact. Each component expectation nevertheless
remains analytically intractable in general because the logarithm
contains the full Gaussian mixture.

For component $i$, let $S_i \in \mathbb{R}^{d\times d}$ be a matrix such that
\begin{equation}
C_i
=
S_iS_i^\top,
\label{eq:covariance_factorization}
\end{equation}
for example, the Cholesky factor of $C_i$. Define
\begin{equation}
X_i
=
\mu_i+S_iY,
\qquad
Y\sim\mathcal{N}(0,I_d),
\label{eq:standardization}
\end{equation}
where $I_d$ denotes the $d\times d$ identity matrix. Since
$Y\sim\mathcal{N}(0,I_d)$ implies
$X_i\sim\mathcal{N}(\mu_i,C_i)$, it follows that
\begin{equation}
\mathbb{E}_{X_i\sim\mathcal{N}(\mu_i,C_i)}
\left[
\log g(X_i)
\right]
=
\mathbb{E}_{Y\sim\mathcal{N}(0,I_d)}
\left[
\log g(\mu_i+S_iY)
\right].
\label{eq:component_expectation_standard}
\end{equation}

For each component, define the standardized integrand
\begin{equation}
f_i(y)
\triangleq
\log g(\mu_i+S_i y).
\label{eq:fi_def}
\end{equation}
Substituting \eqref{eq:fi_def} into \eqref{eq:component_decomp} gives
\begin{equation}
H(g)
=
-\sum_{i=1}^{L}
\omega_i
\mathbb{E}_{Y\sim\mathcal{N}(0,I_d)}
\left[
f_i(Y)
\right].
\label{eq:entropy_as_sum_fi}
\end{equation}
Thus, entropy evaluation reduces to computing a collection of
expectations of the standardized functions $f_i$ under the
$d$-dimensional standard normal distribution.
\section{Taylor Baselines}

For numerical comparison, we consider the Taylor approximations of Huber
\textit{et al.}~\cite{Huber2008MFI}, applied separately to the Gaussian
expectations in \eqref{eq:component_decomp}. Define
\begin{equation}
\ell(x)
\triangleq
\log g(x).
\label{eq:ell_def}
\end{equation}
For component $i$, let
\begin{equation}
X_i \sim \mathcal{N}(\mu_i,C_i).
\end{equation}
A second-order Taylor expansion of $\ell(X_i)$ about the component mean
$\mu_i$ gives
\begin{equation}
\begin{aligned}
\ell(X_i)
\approx\,
&
\ell(\mu_i)
+
\nabla\ell(\mu_i)^\top
(X_i-\mu_i)
\\
&
+
\frac{1}{2}
(X_i-\mu_i)^\top
\nabla^2\ell(\mu_i)
(X_i-\mu_i),
\end{aligned}
\label{eq:taylor_expansion_component}
\end{equation}
where $\nabla\ell(\mu_i)$ and $\nabla^2\ell(\mu_i)$ denote,
respectively, the gradient and Hessian of
$\ell(x)=\log g(x)$ evaluated at $\mu_i$.

Retaining only the constant term gives the zeroth-order approximation
\begin{equation}
H_{\mathrm{T0}}(g)
\triangleq
-\sum_{i=1}^{L}
\omega_i
\ell(\mu_i).
\label{eq:taylor0_general}
\end{equation}
Because
\begin{equation}
\mathbb{E}_{X_i\sim\mathcal{N}(\mu_i,C_i)}
\left[
X_i-\mu_i
\right]
=
0,
\label{eq:first_central_moment}
\end{equation}
the expected first-order Taylor term vanishes. Hence, the zeroth- and
first-order entropy approximations coincide.

Using
\begin{equation}
\mathbb{E}_{X_i\sim\mathcal{N}(\mu_i,C_i)}
\left[
(X_i-\mu_i)(X_i-\mu_i)^\top
\right]
=
C_i,
\label{eq:second_central_moment}
\end{equation}
the expected quadratic term satisfies
\begin{equation}
\mathbb{E}
\left[
(X_i-\mu_i)^\top
\nabla^2\ell(\mu_i)
(X_i-\mu_i)
\right]
=
\operatorname{tr}
\left(
C_i\nabla^2\ell(\mu_i)
\right).
\label{eq:taylor_quadratic_expectation}
\end{equation}
The resulting second-order approximation is therefore
\begin{equation}
H_{\mathrm{T2}}(g)
\triangleq
-\sum_{i=1}^{L}
\omega_i
\left[
\ell(\mu_i)
+
\frac{1}{2}
\operatorname{tr}
\left(
C_i
\nabla^2\ell(\mu_i)
\right)
\right].
\label{eq:taylor2_general}
\end{equation}

Huber \textit{et al.}~\cite{Huber2008MFI} further proposed improving
the local Taylor approximation by replacing a broad Gaussian component
with a finite mixture of narrower Gaussian subcomponents. Specifically,
for each original component $i$, consider the approximation
\begin{equation}
\mathcal{N}(x;\mu_i,C_i)
\approx
\sum_{s=1}^{K_i}
\alpha_{is}\,
\mathcal{N}(x;\mu_{is},C_{is}),
\label{eq:component_splitting}
\end{equation}
where
\begin{equation}
\alpha_{is}\geq 0,
\qquad
\sum_{s=1}^{K_i}\alpha_{is}=1,
\qquad
C_{is}\succ 0.
\label{eq:component_splitting_conditions}
\end{equation}
Here, $K_i$ is the number of subcomponents used to approximate the
$i$th original Gaussian component, $\alpha_{is}$ is the corresponding
conditional split weight, and $\mu_{is}$ and $C_{is}$ are the mean and
covariance of the $s$th subcomponent. The purpose of the splitting is
to represent the probability mass of a broad Gaussian component by
several more localized Gaussian components, thereby improving the
accuracy of the Taylor approximation about the individual subcomponent
means. A component that is not split can be represented by $K_i=1$,
$\alpha_{i1}=1$, $\mu_{i1}=\mu_i$, and $C_{i1}=C_i$.

Applying this replacement to each Gaussian component in the outer
integration density gives
\begin{equation}
\begin{aligned}
\widetilde g_{\mathrm{sp}}(x)
&\triangleq
\sum_{i=1}^{L}
\omega_i
\sum_{s=1}^{K_i}
\alpha_{is}\,
\mathcal{N}(x;\mu_{is},C_{is})
\\
&=
\sum_{r=1}^{L_{\mathrm{sp}}}
\bar{\omega}_r
\mathcal{N}(x;\bar{\mu}_r,\bar{C}_r),
\end{aligned}
\label{eq:split_outer_mixture}
\end{equation}
where
\begin{equation}
L_{\mathrm{sp}}
=
\sum_{i=1}^{L}K_i,
\label{eq:number_split_components}
\end{equation}
and the single index $r$ is used to enumerate all split subcomponents.
The corresponding flattened weights satisfy
\begin{equation}
\bar{\omega}_r\geq 0,
\qquad
\sum_{r=1}^{L_{\mathrm{sp}}}
\bar{\omega}_r=1,
\qquad
\bar{C}_r\succ 0.
\label{eq:split_mixture_conditions}
\end{equation}
Equivalently, each flattened weight is of the form
$\bar{\omega}_r=\omega_i\alpha_{is}$ for the corresponding pair
$(i,s)$.

Importantly, the splitting is applied only to the Gaussian mixture
serving as the outer integration density. Following
\cite{Huber2008MFI}, the original mixture density $g(x)$ is retained
inside the logarithm. Thus, the split approximation replaces
\begin{equation}
-\int_{\mathbb{R}^d} g(x)\log g(x)\,dx
\end{equation}
by
\begin{equation}
-\int_{\mathbb{R}^d}
\widetilde g_{\mathrm{sp}}(x)
\log g(x)\,dx,
\label{eq:split_entropy_integral}
\end{equation}
before applying the local Taylor approximation to $\ell(x)=\log g(x)$.

Applying the second-order Taylor approximation about each split mean
$\bar{\mu}_r$ yields
\begin{equation}
H_{\mathrm{T2,sp}}(g)
\triangleq
-\sum_{r=1}^{L_{\mathrm{sp}}}
\bar{\omega}_r
\left[
\ell(\bar{\mu}_r)
+
\frac{1}{2}
\operatorname{tr}
\left(
\bar{C}_r
\nabla^2\ell(\bar{\mu}_r)
\right)
\right].
\label{eq:taylor2_split}
\end{equation}
The split Taylor baseline therefore contains two distinct sources of
approximation error: the error introduced by representing each selected
Gaussian component with a finite Gaussian mixture and the truncation
error of the second-order Taylor expansion. The splitting rule and the
numbers $K_i$ of split subcomponents used in the numerical experiments
are specified in the corresponding experimental setup.

\section{Hermite Interpretation under the Standard Normal Measure}
\label{sec:hermite_representation}

The entropy evaluator does not require constructing a Hermite expansion.
This section provides an equivalent Hermite interpretation of the
standardized Gaussian expectations and introduces notation that is also used
later for the action-space surrogate.

\subsection{Probabilists' Hermite Basis}

The univariate probabilists' Hermite polynomials
$\{\mathrm{He}_n\}_{n\geq 0}$ satisfy
\begin{equation}
\mathrm{He}_0(y)=1,
\qquad
\mathrm{He}_1(y)=y,
\qquad
\mathrm{He}_{n+1}(y)
=
y\mathrm{He}_n(y)-n\mathrm{He}_{n-1}(y),
\label{eq:hermite_recursion}
\end{equation}
for $n\geq 1$ \cite{AbramowitzStegun1965,Gautschi2004,Xiu2010}.
If $Y\sim\mathcal{N}(0,1)$, then
\begin{equation}
\mathbb{E}
\left[
\mathrm{He}_m(Y)\mathrm{He}_n(Y)
\right]
=
n!\,\delta_{mn},
\qquad
\mathbb{E}
\left[
\mathrm{He}_n(Y)
\right]
=
0
\quad
\text{for } n\geq 1,
\label{eq:hermite_orthogonality_1d}
\end{equation}
where $\delta_{mn}$ is the Kronecker delta.

Let $\mathbb{N}_0=\{0,1,2,\ldots\}$. For a multi-index
$\alpha=(\alpha_1,\ldots,\alpha_d)\in\mathbb{N}_0^d$, define
\begin{equation}
|\alpha|
\triangleq
\sum_{k=1}^{d}\alpha_k,
\qquad
\alpha!
\triangleq
\prod_{k=1}^{d}\alpha_k!,
\label{eq:multi_index_definitions}
\end{equation}
and the multivariate Hermite polynomial
\begin{equation}
\mathrm{He}_{\alpha}(y)
\triangleq
\prod_{k=1}^{d}
\mathrm{He}_{\alpha_k}(y_k),
\qquad
y=(y_1,\ldots,y_d)\in\mathbb{R}^d.
\label{eq:multivariate_hermite}
\end{equation}
For $Y\sim\mathcal{N}(0,I_d)$, these polynomials satisfy
\begin{equation}
\mathbb{E}
\left[
\mathrm{He}_{\alpha}(Y)
\mathrm{He}_{\beta}(Y)
\right]
=
\alpha!\,
\delta_{\alpha\beta},
\label{eq:hermite_orthogonality_dd}
\end{equation}
where $\delta_{\alpha\beta}=1$ if $\alpha=\beta$ and is zero otherwise.
The multivariate probabilists' Hermite polynomials form a complete
orthogonal basis of $L^2(\mathcal{N}(0,I_d))$.

\subsection{Constant-Coefficient Interpretation of the Entropy Contribution}

Under the finite-mixture assumptions in \eqref{eq:gmm}, the standardized
integrand $f_i$ defined in \eqref{eq:fi_def} belongs to
$L^2(\mathcal{N}(0,I_d))$. To establish this property, choose any component
$j_0$ such that $\omega_{j_0}>0$. Since
\begin{equation}
g(x)
\geq
\omega_{j_0}
\mathcal{N}(x;\mu_{j_0},C_{j_0}),
\label{eq:gmm_single_component_lower_bound}
\end{equation}
we have
\begin{equation}
\log g(x)
\geq
\log \omega_{j_0}
+
\log \mathcal{N}(x;\mu_{j_0},C_{j_0}).
\label{eq:log_gmm_lower_bound}
\end{equation}
Because the log-density of a nonsingular Gaussian has the form of a
finite constant minus a positive-definite quadratic form in $x-\mu_{j_0}$,
\eqref{eq:log_gmm_lower_bound} implies that the negative part of
$\log g(x)$ grows at most quadratically in $\|x\|$.

Moreover, each Gaussian component satisfies
\begin{equation}
\mathcal{N}(x;\mu_j,C_j)
\leq
\frac{1}
{(2\pi)^{d/2}\lvert C_j\rvert^{1/2}},
\end{equation}
and hence
\begin{equation}
g(x)
\leq
\sum_{j=1}^{L}
\frac{
\omega_j
}{
(2\pi)^{d/2}
\lvert C_j\rvert^{1/2}
}.
\label{eq:gmm_global_upper_bound}
\end{equation}
Therefore, $\log g(x)$ is bounded above by a finite constant.

Combining the lower and upper bounds, there exist finite constants
$A$ and $B$ such that
\begin{equation}
\left|
\log g(x)
\right|
\leq
A+B\|x\|^2.
\label{eq:log_gmm_quadratic_growth}
\end{equation}
For component $i$, substituting $x=\mu_i+S_i y$ and using the fact that
this transformation is affine in $y$ yields finite constants $A_i$ and
$B_i$ such that
\begin{equation}
\left|
f_i(y)
\right|
=
\left|
\log g(\mu_i+S_i y)
\right|
\leq
A_i+B_i\|y\|^2.
\label{eq:standardized_log_mixture_growth}
\end{equation}
Consequently,
\begin{equation}
f_i(y)^2
\leq
D_i
\left(
1+\|y\|^4
\right)
\label{eq:standardized_log_mixture_square_bound}
\end{equation}
for some finite constant $D_i$. Since a standard Gaussian random vector
has finite fourth moments,
\begin{equation}
\mathbb{E}_{Y\sim\mathcal{N}(0,I_d)}
\left[
f_i(Y)^2
\right]
<
\infty.
\label{eq:standardized_log_mixture_square_integrable}
\end{equation}
Therefore,
\begin{equation}
f_i
\in
L^2(\mathcal{N}(0,I_d)).
\end{equation}

Because the multivariate Hermite polynomials form a complete orthogonal
basis of $L^2(\mathcal{N}(0,I_d))$, the function $f_i$ admits the Hermite
expansion
\begin{equation}
f_i
=
\sum_{\alpha\in\mathbb{N}_0^d}
c_{i,\alpha}
\mathrm{He}_{\alpha}
\qquad
\text{in }L^2(\mathcal{N}(0,I_d)),
\label{eq:hermite_representation}
\end{equation}
with coefficients
\begin{equation}
c_{i,\alpha}
=
\frac{1}{\alpha!}
\mathbb{E}_{Y\sim\mathcal{N}(0,I_d)}
\left[
f_i(Y)
\mathrm{He}_{\alpha}(Y)
\right].
\label{eq:hermite_coefficients}
\end{equation}
Thus, the series in \eqref{eq:hermite_representation} converges to
$f_i$ in the $L^2(\mathcal{N}(0,I_d))$ sense.

Let $\boldsymbol{0}=(0,\ldots,0)$ denote the zero multi-index. Since
\begin{equation}
\boldsymbol{0}!=1,
\qquad
\mathrm{He}_{\boldsymbol{0}}(y)=1,
\end{equation}
the coefficient formula \eqref{eq:hermite_coefficients} gives
\begin{equation}
c_{i,\boldsymbol{0}}
=
\mathbb{E}_{Y\sim\mathcal{N}(0,I_d)}
\left[
f_i(Y)
\right].
\label{eq:constant_hermite_coefficient}
\end{equation}
Substituting \eqref{eq:constant_hermite_coefficient} into
\eqref{eq:entropy_as_sum_fi} gives the exact identity
\begin{equation}
H(g)
=
-\sum_{i=1}^{L}
\omega_i
c_{i,\boldsymbol{0}}.
\label{eq:entropy_from_constant_coefficient}
\end{equation}

Equation~\eqref{eq:constant_hermite_coefficient} shows that each standardized
component expectation equals the constant Hermite coefficient of $f_i$.
This identity is interpretive rather than an additional numerical
approximation. Entropy evaluation does not require constructing a truncated
Hermite expansion or computing higher-order Hermite coefficients; the
Gaussian expectation itself is evaluated directly by Gauss--Hermite
quadrature in the next section.

\section{Gauss--Hermite Entropy Evaluation}
\label{sec:gh_evaluation}

The standardized Gaussian expectation in
\eqref{eq:component_expectation_standard}, equivalently the constant Hermite
coefficient in \eqref{eq:constant_hermite_coefficient}, is evaluated directly
by Gauss--Hermite quadrature. For a $d$-dimensional standard normal random
vector $Y\sim\mathcal{N}(0,I_d)$, let
$\{(T_m,W_m)\}_{m=1}^{M}$ denote a Gauss--Hermite rule associated with
the weight $\exp(-\|t\|^2)$. Then
\begin{equation}
\mathbb{E}_{Y\sim\mathcal{N}(0,I_d)}
\left[
h(Y)
\right]
\approx
\frac{1}{\pi^{d/2}}
\sum_{m=1}^{M}
W_m
h(\sqrt{2}\,T_m).
\label{eq:gh_standard_normal_dd}
\end{equation}
The factors $\sqrt{2}$ and $\pi^{-d/2}$ convert the conventional
Gauss--Hermite weight $\exp(-\|t\|^2)$ to the standard normal measure.
For the full tensor-product rule used here, with $Q$ nodes in each
dimension,
\begin{equation}
M=Q^d.
\label{eq:tensor_node_count}
\end{equation}
The one-dimensional nodes and weights underlying the tensor-product
construction are the standard Gauss--Hermite nodes and weights
\cite{AbramowitzStegun1965,Gautschi2004,Xiu2010}.

Applying \eqref{eq:gh_standard_normal_dd} to
\eqref{eq:constant_hermite_coefficient} gives
\begin{equation}
\widehat{c}_{i,\boldsymbol{0}}^{(Q)}
=
\frac{1}{\pi^{d/2}}
\sum_{m=1}^{M}
W_m
\log g
\left(
\mu_i+\sqrt{2}\,S_iT_m
\right).
\label{eq:constant_coefficient_gh}
\end{equation}
Substitution into \eqref{eq:entropy_from_constant_coefficient} yields
the Gauss--Hermite entropy estimator
\begin{equation}
\boxed{
\widehat{H}_{Q}(g)
=
-\frac{1}{\pi^{d/2}}
\sum_{i=1}^{L}
\omega_i
\sum_{m=1}^{M}
W_m
\log g
\left(
\mu_i+\sqrt{2}\,S_iT_m
\right)
}.
\label{eq:entropy_gh_estimator}
\end{equation}
Thus, the original Gaussian-mixture entropy integral is replaced by
$L$ standardized Gaussian expectations, each evaluated at the same
Gauss--Hermite nodes. The subscript $Q$ denotes the quadrature order per
dimension; it is unrelated to a truncation order of the Hermite
representation.

For numerical stability, $\log g(x)$ in
\eqref{eq:entropy_gh_estimator} is evaluated in log-sum-exp form. At
quadrature node
\begin{equation}
x_{i,m}
\triangleq
\mu_i+\sqrt{2}\,S_iT_m,
\label{eq:quadrature_state_node}
\end{equation}
assume that zero-weight mixture components have been removed and define
\begin{equation}
a_j(x)
\triangleq
\log\omega_j
+
\log\mathcal{N}(x;\mu_j,C_j),
\qquad
j=1,\ldots,L,
\label{eq:log_component_term}
\end{equation}
and
\begin{equation}
a_{\max}(x)
\triangleq
\max_{1\leq j\leq L}a_j(x).
\label{eq:maximum_log_component}
\end{equation}
Then
\begin{equation}
\log g(x)
=
a_{\max}(x)
+
\log
\left[
\sum_{j=1}^{L}
\exp
\left(
a_j(x)-a_{\max}(x)
\right)
\right].
\label{eq:stable_logsumexp}
\end{equation}
This form avoids direct evaluation of very small weighted Gaussian
densities and thereby reduces numerical underflow in the tails of the
mixture.

The computational cost follows directly from
\eqref{eq:entropy_gh_estimator}. For each outer component $i$ and
quadrature node $m$, evaluating $\log g(x_{i,m})$ requires evaluating
all $L$ Gaussian component log-densities. With dense covariance
matrices and precomputed covariance factorizations, one Gaussian
log-density evaluation requires $O(d^2)$ operations. The dominant cost
is therefore
\begin{equation}
O\left(L^2Md^2\right)
=
O\left(L^2Q^d d^2\right)
\label{eq:entropy_cost}
\end{equation}
for the full tensor-product rule. Computing the covariance
factorizations
\begin{equation}
C_i=S_iS_i^\top,
\qquad i=1,\ldots,L,
\end{equation}
requires a one-time cost of
\begin{equation}
O\left(Ld^3\right).
\label{eq:cholesky_cost}
\end{equation}
This preprocessing cost can be amortized when the same covariance
matrices are reused across repeated entropy evaluations.

For entropy-only computation, $Q$ is therefore the numerical resolution
parameter. No truncated Hermite expansion or higher-order Hermite
coefficients are constructed. In practice, quadrature adequacy can be
assessed by increasing $Q$ and comparing successive entropy estimates,
or by comparison with an independent numerical reference when one is
available; such comparisons provide convergence diagnostics rather than
rigorous error bounds.

\section{One-Dimensional Parzen-GMM Entropy Benchmark}
\label{sec:num_1d_huber}

We consider the one-dimensional parameter-identification example of
Huber et al.~\cite{Huber2008MFI} as a benchmark for deterministic
entropy evaluation of a data-driven Gaussian mixture. The latent random variable $X$ has density
\begin{equation}
g_x(x)
\triangleq
0.4\,\mathcal{N}(x;-1,0.025)
+
0.6\,\mathcal{N}(x;1,1),
\label{eq:huber_x}
\end{equation}
and the measurement model is
\begin{equation}
Y
=
aX+W,
\qquad
a=2,
\label{eq:huber_y}
\end{equation}
where $W$ is a zero-mean Gaussian random variable independent of $X$.

\subsection{Residual Definition for Replication}

Huber et al.~\cite{Huber2008MFI} define a scaled quantity
$\widetilde{Y}=\widetilde{a}Y$ and write the residual as
$E=Y-\widetilde{Y}$, while also stating that the residual entropy is
minimized at the inverse coefficient $\widetilde{a}=1/a$. Taken
literally, those definitions give
\begin{equation}
E
=
(1-\widetilde{a})Y,
\label{eq:huber_literal_residual}
\end{equation}
which becomes degenerate at $\widetilde{a}=1$, rather than at
$\widetilde{a}=1/a$. The stated inverse coefficient and the minimum
shown in Fig.~3 of \cite{Huber2008MFI} are therefore not consistent
with the residual definition printed in the original paper.

For the replication, we adopt the estimator
\begin{equation}
\widehat{X}(\widetilde{a})
\triangleq
\widetilde{a}Y
\label{eq:huber_x_estimate}
\end{equation}
and define
\begin{equation}
\begin{aligned}
E(\widetilde{a})
&\triangleq
X-\widehat{X}(\widetilde{a})
\\
&=
X-\widetilde{a}(aX+W)
\\
&=
(1-a\widetilde{a})X-\widetilde{a}W.
\end{aligned}
\label{eq:huber_e}
\end{equation}
This residual definition is used throughout the present benchmark.

The algebraic inverse
\begin{equation}
\widetilde{a}
=
\frac{1}{a}
\label{eq:algebraic_inverse}
\end{equation}
is a distinguished reference point because it eliminates the
input-dependent term:
\begin{equation}
E(1/a)
=
-\frac{1}{a}W.
\label{eq:residual_at_inverse}
\end{equation}
This cancellation property does not, however, imply that $1/a$
generally minimizes the residual entropy. Changing $\widetilde{a}$
changes both the residual input term and the effective noise scale.

For example, suppose that $X$ and $W$ are independent Gaussian random
variables with variances $\sigma_x^2$ and $\sigma_w^2$, respectively.
Then
\begin{equation}
\operatorname{Var}
\left[
E(\widetilde{a})
\right]
=
(1-a\widetilde{a})^2\sigma_x^2
+
\widetilde{a}^{\,2}\sigma_w^2.
\label{eq:gaussian_residual_variance}
\end{equation}
Because the entropy of a nondegenerate Gaussian random variable is
strictly increasing in its variance, the minimizing coefficient is
\begin{equation}
\widetilde{a}_{\mathrm{ent}}
=
\frac{a\sigma_x^2}
{a^2\sigma_x^2+\sigma_w^2}.
\label{eq:gaussian_entropy_minimizer}
\end{equation}
Thus,
\begin{equation}
\widetilde{a}_{\mathrm{ent}}
\neq
\frac{1}{a}
\qquad
\text{when }
\sigma_w^2>0.
\label{eq:inverse_not_entropy_minimizer}
\end{equation}
For scale, the variance of the non-Gaussian latent variable in
\eqref{eq:huber_x} is
\begin{equation}
\operatorname{Var}(X)=1.57.
\end{equation}
If $X$ is replaced only for this diagnostic calculation by a
moment-matched Gaussian with variance $1.57$,
\eqref{eq:gaussian_entropy_minimizer} gives
\begin{equation}
\widetilde a_{\mathrm{ent}}\approx0.49968
\qquad\text{for }\sigma_w^2=0.004,
\end{equation}
and
\begin{equation}
\widetilde a_{\mathrm{ent}}\approx0.49684
\qquad\text{for }\sigma_w^2=0.04.
\end{equation}
Thus, although the algebraic inverse and the Gaussian entropy minimizer are
not identical in general, they are numerically very close for a moment-matched
Gaussian with the parameters of this benchmark. This calculation illustrates
the distinction only; it is not an assertion about the exact minimizer of the
non-Gaussian finite-sample Parzen-GMM entropy.

The present experiment is therefore used to compare entropy-evaluation
methods over the parameter interval
\begin{equation}
\widetilde{a}
\in
[-2,6],
\label{eq:huber_parameter_interval}
\end{equation}
rather than to assert that minimum residual entropy necessarily
identifies the exact inverse coefficient.

\subsection{Parzen-GMM Construction}

Following Huber et al.~\cite{Huber2008MFI}, we evaluate a Parzen
estimate~\cite{Parzen1962} of the residual density rather than the analytic density
induced directly by \eqref{eq:huber_x}--\eqref{eq:huber_e}. A single
set of independent sample pairs
\begin{equation}
\left\{
(X_n,W_n)
\right\}_{n=1}^{N}
\label{eq:huber_sample_pairs}
\end{equation}
is generated once and reused for every candidate value of
$\widetilde{a}$. The same fixed sample realization is used for all
entropy-evaluation methods in this benchmark. The residual samples are
\begin{equation}
E_n(\widetilde{a})
=
(1-a\widetilde{a})X_n
-
\widetilde{a}W_n,
\qquad
n=1,\ldots,N.
\label{eq:huber_residual_samples}
\end{equation}

We use the same sample count as the original study,
\begin{equation}
N
=
100.
\label{eq:huber_sample_count}
\end{equation}
For each $\widetilde{a}$, the residual density is approximated using
the Gaussian-kernel Parzen estimator
\begin{equation}
\widehat{g}_{\widetilde{a}}(e)
\triangleq
\frac{1}{N}
\sum_{n=1}^{N}
\mathcal{N}
\left(
e;
E_n(\widetilde{a}),
h(\widetilde{a})^2
\right).
\label{eq:parzen_gmm}
\end{equation}
Thus, $\widehat{g}_{\widetilde{a}}$ is an $N$-component equal-weight
Gaussian mixture.

The kernel bandwidth is selected using Silverman's rule of thumb~\cite{Silverman1986}:
\begin{equation}
h(\widetilde{a})
=
1.06\,
\widehat{\sigma}_E(\widetilde{a})
N^{-1/5},
\label{eq:silverman}
\end{equation}
where $\widehat{\sigma}_E(\widetilde{a})$ is the sample standard
deviation of
$\{E_n(\widetilde{a})\}_{n=1}^{N}$.

Because differential entropy is concave in the density, the equal-weight
Gaussian mixture in \eqref{eq:parzen_gmm} satisfies the sample-dependent
lower bound
\begin{align}
H\!\left(\widehat g_{\widetilde a}\right)
&\geq
\frac{1}{N}\sum_{n=1}^{N}
H\!\left(\mathcal{N}(E_n(\widetilde a),h(\widetilde a)^2)\right)
\nonumber\\
&=\frac{1}{2}\log\!\left(2\pi e\,h(\widetilde a)^2\right).
\label{eq:parzen_entropy_bandwidth_lower_bound}
\end{align}
This bound depends on the realized sample through $h(\widetilde a)$ and
therefore provides a direct consistency check for a specified sample
realization and bandwidth.

The residual family also has an explicit scale structure. Define
\begin{equation}
s_{\widetilde a}\triangleq1-a\widetilde a,
\qquad
\beta_{\widetilde a}\triangleq\frac{\widetilde a}{s_{\widetilde a}},
\qquad s_{\widetilde a}\neq0.
\end{equation}
Then
\begin{equation}
E_n(\widetilde a)=s_{\widetilde a}\left(X_n-\beta_{\widetilde a}W_n\right).
\end{equation}
Let $\widehat g_{\beta}^{(V)}$ denote the Gaussian-kernel Parzen density
formed from the samples $V_{n,\beta}=X_n-\beta W_n$ using the same
Silverman rule, and let $h_{\beta}^{(V)}$ be its bandwidth. Scale
equivariance of the sample standard deviation gives
\begin{equation}
h(\widetilde a)=|s_{\widetilde a}|\,h_{\beta_{\widetilde a}}^{(V)}.
\end{equation}
Consequently,
\begin{equation}
\widehat g_{\widetilde a}(e)
=
\frac{1}{|s_{\widetilde a}|}
\widehat g_{\beta_{\widetilde a}}^{(V)}\!\left(\frac{e}{s_{\widetilde a}}\right),
\end{equation}
and hence
\begin{equation}
H\!\left(\widehat g_{\widetilde a}\right)
=
H\!\left(\widehat g_{\beta_{\widetilde a}}^{(V)}\right)
+\log|s_{\widetilde a}|.
\label{eq:parzen_scale_structure}
\end{equation}
The second term isolates the explicit entropy shift caused by scaling. The
first term is not constant because $\beta_{\widetilde a}$ also varies with
$\widetilde a$; the sweep is therefore not merely a collection of exact scale
copies of one fixed density. The singular case $s_{\widetilde a}=0$ is
evaluated directly from \eqref{eq:huber_residual_samples}.

The quantity evaluated below is the differential entropy
\begin{equation}
H
\left(
\widehat{g}_{\widetilde{a}}
\right),
\label{eq:parzen_entropy_target}
\end{equation}
not the entropy of the analytic residual distribution itself.

\subsection{Entropy Evaluation}

Using the entropy decomposition in
\eqref{eq:component_decomp}, the entropy of the Parzen-GMM is
\begin{equation}
H
\left(
\widehat{g}_{\widetilde{a}}
\right)
=
-\frac{1}{N}
\sum_{i=1}^{N}
\mathbb{E}_{
Z_i
\sim
\mathcal{N}
\left(
E_i(\widetilde{a}),
h(\widetilde{a})^2
\right)
}
\left[
\log
\widehat{g}_{\widetilde{a}}(Z_i)
\right].
\label{eq:parzen_entropy_decomp}
\end{equation}

For this one-dimensional equal-weight mixture, the zeroth- and
second-order Taylor baselines are obtained from
\eqref{eq:taylor0_general} and \eqref{eq:taylor2_general} by setting
\begin{equation}
L=N,
\qquad
\omega_i=\frac{1}{N},
\qquad
\mu_i=E_i(\widetilde{a}),
\qquad
C_i=h(\widetilde{a})^2.
\label{eq:parzen_taylor_substitution}
\end{equation}

For the proposed method, introduce the standard normal
reparameterization
\begin{equation}
Z_i
=
E_i(\widetilde{a})
+
h(\widetilde{a})U,
\qquad
U\sim\mathcal{N}(0,1).
\label{eq:parzen_standardization}
\end{equation}
Applying the one-dimensional specialization of
\eqref{eq:entropy_gh_estimator} gives
\begin{equation}
\widehat{H}_{Q}(\widetilde{a})
\triangleq
-\frac{1}{N\sqrt{\pi}}
\sum_{i=1}^{N}
\sum_{q=1}^{Q}
w_q
\log
\widehat{g}_{\widetilde{a}}
\left(
E_i(\widetilde{a})
+
\sqrt{2}\,
h(\widetilde{a})t_q
\right).
\label{eq:parzen_HQ}
\end{equation}
At every quadrature point, the logarithm of the Parzen-GMM density is
evaluated using the numerically stable expression in
\eqref{eq:stable_logsumexp}.

\subsection{Replication Settings}

Huber et al.~\cite{Huber2008MFI} report the measurement-noise variance
as
\begin{equation}
\sigma_w^2
=
0.04.
\label{eq:huber_reported_noise_variance}
\end{equation}
Under the residual interpretation in \eqref{eq:huber_e}, the
input-dependent term vanishes at $\widetilde{a}=1/a=0.5$. The
corresponding analytic residual is Gaussian and has entropy
\begin{equation}
H
\left(
E(1/a)
\right)
=
\frac{1}{2}
\log
\left(
2\pi e\,
\frac{\sigma_w^2}{a^2}
\right).
\label{eq:gauss_inverse_entropy}
\end{equation}
For the reported variance $\sigma_w^2=0.04$,
\eqref{eq:gauss_inverse_entropy} gives
\begin{equation}
H
\left(
E(0.5)
\right)
\approx
-0.884
\quad
\text{nats}.
\label{eq:inverse_entropy_reported_variance}
\end{equation}
In contrast, using
\begin{equation}
\sigma_w^2
=
0.004
\label{eq:huber_replication_noise_variance}
\end{equation}
gives
\begin{equation}
H
\left(
E(0.5)
\right)
\approx
-2.035
\quad
\text{nats},
\label{eq:inverse_entropy_replication_variance}
\end{equation}
which is substantially closer to the entropy scale near the minimum
shown in Fig.~3 of \cite{Huber2008MFI}.

The values in \eqref{eq:inverse_entropy_reported_variance} and
\eqref{eq:inverse_entropy_replication_variance} are analytic
Gaussian-residual entropies and are used only as scale diagnostics.
They are not identical to the entropy of the finite-sample
Parzen-GMM, which also depends on the sample realization and kernel
bandwidth. Because the original paper does not provide the Parzen bandwidth, the
measurement-noise variance underlying the published entropy curve cannot be
established conclusively from that curve alone; the finite-sample Parzen curve
also depends on the realized sample. Equation~\eqref{eq:parzen_entropy_bandwidth_lower_bound}
provides a sample-specific consistency check whenever the bandwidth and sample
realization are specified.

We use $\sigma_w^2=0.004$ as the replication setting that reproduces the
reported entropy scale more closely. The source-reported value
$\sigma_w^2=0.04$ is retained separately in the scale comparison above.

\subsection{Reference Computation and Results}

For each value of $\widetilde{a}$, the numerical reference is obtained
by independent numerical integration,
\begin{equation}
H_{\mathrm{ref}}(\widetilde{a})
\triangleq
-
\int_{\mathbb{R}}
\widehat{g}_{\widetilde{a}}(e)
\log
\widehat{g}_{\widetilde{a}}(e)
\,de.
\label{eq:parzen_reference_entropy}
\end{equation}

The numerical reference is compared with the zeroth- and second-order
Taylor approximations in \eqref{eq:taylor0_general} and
\eqref{eq:taylor2_general}, respectively, and with the proposed
Gauss--Hermite estimator \eqref{eq:parzen_HQ} for
\begin{equation}
Q
\in
\{3,5\}.
\label{eq:one_dimensional_gh_orders}
\end{equation}

For an entropy-evaluation method $\mathcal{M}$, define the error
\begin{equation}
\Delta H_{\mathcal{M}}(\widetilde{a})
\triangleq
\widehat{H}_{\mathcal{M}}(\widetilde{a})
-
H_{\mathrm{ref}}(\widetilde{a}),
\label{eq:entropy_error_1d}
\end{equation}
where
\begin{equation}
\mathcal{M}
\in
\left\{
\mathrm{T0},
\mathrm{T2},
\mathrm{GH3},
\mathrm{GH5}
\right\}.
\label{eq:one_dimensional_methods}
\end{equation}

\begin{figure}[t]
    \centering
    \includegraphics[width=0.92\linewidth]{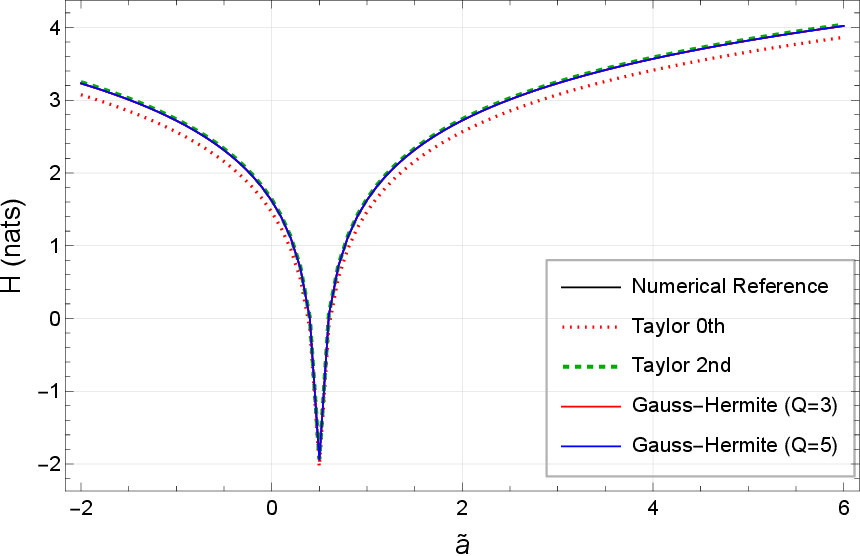}
    \caption{
    Entropy of the Parzen-GMM as a function of the candidate inverse
    coefficient $\widetilde{a}$. The numerical reference is compared
    with the zeroth- and second-order Taylor approximations and the proposed
    Gauss--Hermite estimator \eqref{eq:parzen_HQ} for
    $Q\in\{3,5\}$.
    }
    \label{fig:univariate}
\end{figure}

\begin{figure}[t]
    \centering
    \includegraphics[width=0.92\linewidth]{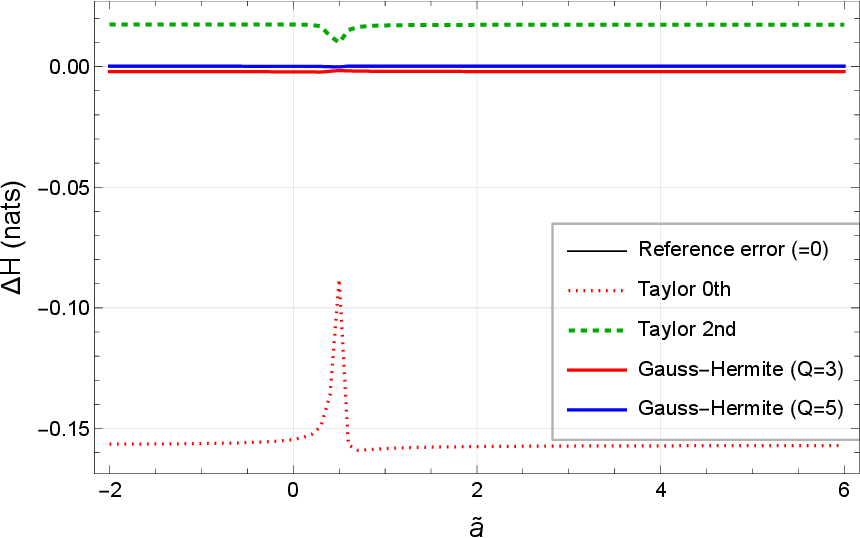}
    \caption{
    Signed entropy-estimation error relative to the numerical reference,
    $\Delta H_{\mathcal{M}}(\widetilde{a})$, for the Taylor-0,
    second-order Taylor approximation, GH3, and GH5 methods. 
    }
    \label{fig:univariateDelta}
\end{figure}

Figure~\ref{fig:univariate} compares the estimated entropy curves over
the parameter sweep, and Figure~\ref{fig:univariateDelta} shows the
corresponding signed errors relative to the numerical reference. At
the scale of the error plot, GH5 remains nearly coincident with the
zero-reference line, while GH3 also exhibits only small deviations.
The Taylor-0 approximation shows a pronounced negative bias, whereas
the second-order Taylor approximation has a smaller positive bias.
Equation~\eqref{eq:parzen_scale_structure} shows that a substantial part of
the vertical variation in this benchmark is an explicit scale effect. The
one-dimensional sweep primarily probes the resulting one-parameter residual
family, while the two-dimensional benchmark below introduces independent
variation in component overlap and separation.

\section{Two-Dimensional Gaussian-Mixture Benchmark}
\label{sec:benchmark_multivariate}

We consider the two-dimensional five-component Gaussian-mixture family used by
Huber et al.~\cite{Huber2008MFI}. The purpose of this benchmark is to compare
the proposed Gauss--Hermite entropy estimator with analytic entropy bounds and
Taylor approximations about the means of the Gaussian components on a mixture
whose degree of overlap and multimodality changes continuously with a scalar parameter.

\subsection{Mixture Family}

The mixture has uniform weights
\begin{equation}
\omega_i
=
\frac{1}{5},
\qquad
i=1,\ldots,5,
\label{eq:multivariate_weights}
\end{equation}
and component means
\begin{equation}
\begin{aligned}
\mu_1(c)
&=
\begin{bmatrix}
0\\
0
\end{bmatrix},
&
\mu_2(c)
&=
\begin{bmatrix}
3\\
2
\end{bmatrix},
&
\mu_3(c)
&=
\begin{bmatrix}
1\\
-0.5
\end{bmatrix},
\\[0.5em]
\mu_4(c)
&=
\begin{bmatrix}
2.5\\
1.5
\end{bmatrix},
&
\mu_5(c)
&=
c
\begin{bmatrix}
1\\
1
\end{bmatrix},
&
c
&\in[-3,3].
\end{aligned}
\label{eq:multivariate_means}
\end{equation}
Only the fifth component mean varies with $c$; the argument $c$ is retained
for $\mu_1,\ldots,\mu_4$ only to keep a uniform notation for the
five-component family.
The covariance matrices are
\begin{equation}
\begin{aligned}
C_1
&=
\operatorname{diag}(0.16,1),
&
C_2
&=
\operatorname{diag}(1,0.16),
\\
C_3
&=
C_4
=
C_5
=
\operatorname{diag}(0.5,0.5).
\end{aligned}
\label{eq:multivariate_covs}
\end{equation}
For each value of $c$, the corresponding density is
\begin{equation}
g_c(x)
\triangleq
\sum_{i=1}^{5}
\omega_i
\mathcal{N}\!\left(x;\mu_i(c),C_i\right),
\qquad
x\in\mathbb{R}^2,
\label{eq:multivariate_density}
\end{equation}
and the target entropy is
\begin{equation}
H(g_c)
\triangleq
-
\int_{\mathbb{R}^2}
g_c(x)
\log g_c(x)
\,dx.
\label{eq:multivariate_entropy}
\end{equation}

\subsection{Reference and Comparison Methods}

\subsubsection{Numerical reference}

The reference entropy is obtained by independent adaptive numerical integration
of \eqref{eq:multivariate_entropy}:
\begin{equation}
H_{\mathrm{ref}}(c)
\triangleq
-
\int_{\mathbb{R}^2}
g_c(x)
\log g_c(x)
\,dx.
\label{eq:multivariate_reference_entropy}
\end{equation}
The logarithm of the Gaussian-mixture density is evaluated using the stable
expression in \eqref{eq:stable_logsumexp}. As an independent numerical
cross-check, the reference calculation is compared with a tensor-product
Gauss--Hermite calculation using $Q_{\mathrm{ref}}=41$. Over the parameter
grid, the maximum absolute discrepancy between the two calculations is
$9.84\times10^{-6}$ nats. The independently integrated values are used as
the reference curve.

\subsubsection{Gauss--Hermite estimator}

For each covariance matrix, let
\begin{equation}
C_i
=
S_iS_i^{\top}
\label{eq:multivariate_cov_factor}
\end{equation}
be its Cholesky factorization. Applying the two-dimensional
specialization of \eqref{eq:entropy_gh_estimator} gives
\begin{equation}
\begin{aligned}
\widehat{H}_{Q}(g_c)
=
-
\frac{1}{5\pi}
\sum_{i=1}^{5}
\sum_{q_1=1}^{Q}
\sum_{q_2=1}^{Q}
&w_{q_1}w_{q_2}
\\
{}
&\times
\log g_c
\left(
\mu_i(c)
+
S_i\sqrt{2}
\begin{bmatrix}
t_{q_1}\\
t_{q_2}
\end{bmatrix}
\right),
\end{aligned}
\label{eq:multivariate_gh_estimator}
\end{equation}
where $\{(t_q,w_q)\}_{q=1}^{Q}$ are the univariate Gauss--Hermite nodes
and weights for the weight $\exp(-t^2)$. The full tensor-product rule uses
$Q^2$ nodes for each outer Gaussian component.

\subsubsection{Analytic lower and upper bounds}

Applying Jensen's inequality to the convex function $-\log x$ gives the lower
bound of Huber et al.~\cite{Huber2008MFI}:
\begin{equation}
H_{\mathrm{LB}}(g_c)
\triangleq
-
\sum_{i=1}^{5}
\omega_i
\log
\left[
\sum_{j=1}^{5}
\omega_j
\mathcal{N}
\left(
\mu_i(c);
\mu_j(c),
C_i+C_j
\right)
\right].
\label{eq:multivariate_lower_bound}
\end{equation}
The corresponding basic component upper bound is
\begin{equation}
H_{\mathrm{UB}}^{\mathrm{basic}}(g_c)
\triangleq
\sum_{i=1}^{5}
\omega_i
\left[
-\log\omega_i
+
\frac{1}{2}
\log
\left(
(2\pi e)^2
\lvert C_i\rvert
\right)
\right].
\label{eq:multivariate_basic_upper_bound}
\end{equation}

To construct the refined upper bound, Gaussian components are merged
successively using moment-preserving pairwise reduction. Following
Huber et al.~\cite{Huber2008MFI}, the pair selected at each step is determined
using the reduction criterion of Runnalls~\cite{Runnalls2007}. Let
\begin{equation}
g_c^{(r)}(x)
=
\sum_{\ell=1}^{L_r}
\omega_{\ell}^{(r)}
\mathcal{N}
\left(
x;
\mu_{\ell}^{(r)}(c),
C_{\ell}^{(r)}(c)
\right),
\qquad
L_r=5-r,
\label{eq:multivariate_reduced_mixture}
\end{equation}
denote the mixture after $r$ pairwise merging operations, with
$g_c^{(0)}=g_c$. For each reduced mixture, define
\begin{equation}
H_{\mathrm{UB}}^{(r)}(g_c)
\triangleq
\sum_{\ell=1}^{L_r}
\omega_{\ell}^{(r)}
\left[
-\log\omega_{\ell}^{(r)}
+
\frac{1}{2}
\log
\left(
(2\pi e)^2
\left\lvert
C_{\ell}^{(r)}(c)
\right\rvert
\right)
\right].
\label{eq:multivariate_reduced_upper_bound}
\end{equation}
Following \cite{Huber2008MFI}, each moment-preserving reduction yields
an entropy upper bound for the original mixture $g_c$. Hence,
$H_{\mathrm{UB}}^{(r)}(g_c)$ remains an upper bound on $H(g_c)$ at
each stage of the merging sequence.
The refined upper bound is the smallest value encountered along the merging
sequence:
\begin{equation}
H_{\mathrm{UB}}^{\mathrm{ref}}(g_c)
\triangleq
\min_{0\leq r\leq 4}
H_{\mathrm{UB}}^{(r)}(g_c).
\label{eq:multivariate_refined_upper_bound}
\end{equation}

For comparison, we also use the entropy of the single Gaussian that matches
the first two moments of $g_c$. Its mean and covariance are
\begin{equation}
\overline{\mu}(c)
\triangleq
\sum_{i=1}^{5}
\omega_i
\mu_i(c),
\label{eq:multivariate_moment_mean}
\end{equation}
and
\begin{equation}
\overline{C}(c)
\triangleq
\sum_{i=1}^{5}
\omega_i
\left[
C_i
+
\left(
\mu_i(c)-\overline{\mu}(c)
\right)
\left(
\mu_i(c)-\overline{\mu}(c)
\right)^{\top}
\right].
\label{eq:multivariate_moment_covariance}
\end{equation}
Because the Gaussian distribution maximizes differential entropy among
distributions with a fixed covariance, the Gaussian matching the first
two moments of $g_c$ provides the upper bound
\begin{equation}
H_{\mathrm{SG}}(g_c)
\triangleq
\frac{1}{2}
\log
\left(
(2\pi e)^2
\left\lvert
\overline{C}(c)
\right\rvert
\right).
\label{eq:multivariate_single_gaussian_bound}
\end{equation}
The endpoint cases of the reduction sequence recover the two upper bounds
introduced above. In particular,
\begin{equation}
H_{\mathrm{UB}}^{(0)}(g_c)=H_{\mathrm{UB}}^{\mathrm{basic}}(g_c).
\end{equation}
Because the pairwise merging is moment preserving, after four merges the
single remaining Gaussian has mean $\overline{\mu}(c)$ and covariance
$\overline{C}(c)$. Hence,
\begin{equation}
H_{\mathrm{UB}}^{(4)}(g_c)=H_{\mathrm{SG}}(g_c).
\end{equation}
It follows directly from \eqref{eq:multivariate_refined_upper_bound} that
\begin{equation}
H_{\mathrm{UB}}^{\mathrm{ref}}(g_c)
\leq
\min\!\left\{H_{\mathrm{UB}}^{\mathrm{basic}}(g_c),H_{\mathrm{SG}}(g_c)\right\}.
\end{equation}
Thus, the refined bound is no larger than the single-Gaussian bound by
construction rather than as an empirical consequence of the benchmark.

\subsubsection{Taylor baselines}

The unsplit second-order Taylor baseline is evaluated using
\eqref{eq:taylor2_general}. The split second-order Taylor baseline is evaluated using
\eqref{eq:taylor2_split}. For the split baseline, we follow the multivariate
setting of Huber et al.~\cite{Huber2008MFI}: at each operation, the component
whose covariance has the largest principal-axis variance is split along that
principal axis using their four-component splitting library, and at most
20 splitting operations are permitted. Splitting modifies only the Gaussian
mixture outside the logarithm; the original density $g_c$ is retained inside
the logarithm. Starting from five components, at most 20 four-way splitting
operations produce at most
\begin{equation}
L_{\mathrm{sp}}\leq5+20(4-1)=65
\end{equation}
outer Gaussian subcomponents. The split Taylor calculation therefore requires
values of $\ell$ and its Hessian at as many as 65 split means, whereas the
Gauss--Hermite entropy estimator uses $5Q^2$ evaluations of
$\log g_c(x)$ in this two-dimensional benchmark. Measured runtime or an
implementation-level primitive-operation count provides the corresponding
implementation-level cost comparison.

All bound and Taylor curves are recomputed from these definitions. Curves
digitized from Fig.~4 of \cite{Huber2008MFI} are not used as numerical data.

\subsection{Numerical Settings}

The continuation parameter $c$ is evaluated on a uniform finite grid
over the interval $[-3,3]$.
The proposed estimator is evaluated at
\begin{equation}
Q
\in
\{3,5\}.
\label{eq:multivariate_reported_orders}
\end{equation}
Gaussian component terms and $\log g_c(x)$ are evaluated in the log domain. The same parameter grid is used for every method.

For an approximation method $\mathcal{M}$, define the signed entropy-estimation error
\begin{equation}
\Delta H_{\mathcal M}(c)
\triangleq
\widehat H_{\mathcal M}(c)
-
H_{\mathrm{ref}}(c).
\label{eq:multivariate_entropy_error}
\end{equation}
The error comparison includes
\begin{equation}
\mathcal{M}
\in
\left\{
\mathrm{T2},
\mathrm{T2,sp},
\mathrm{GH3},
\mathrm{GH5}
\right\}.
\label{eq:multivariate_error_methods}
\end{equation}
The analytic bounds are omitted from the error plot because their purpose is
to bracket the entropy rather than to approximate it symmetrically.

\subsection{Results}

\begin{figure}[t]
    \centering
    \includegraphics[width=\linewidth]{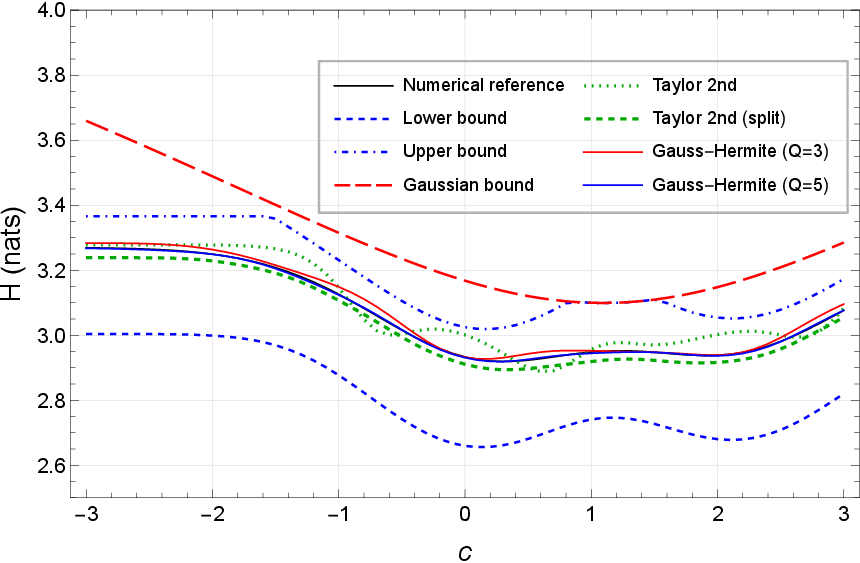}
    \caption{
    Entropy estimates for the two-dimensional five-component
    Gaussian-mixture family. The independently integrated numerical reference
    is compared with the Jensen lower bound, the refined upper bound, the
    single-Gaussian upper bound, the second-order Taylor approximations with and without
    splitting, and the proposed Gauss--Hermite estimator for
    $Q\in\{3,5\}$.
    }
    \label{fig:multivariate}
\end{figure}

Figure~\ref{fig:multivariate} compares the numerical reference with the Jensen
lower bound, the refined upper bound, the single-Gaussian upper bound, the
second-order Taylor approximations with and without component splitting, and the proposed
Gauss--Hermite estimates. The recomputed bound curves reproduce the qualitative behavior reported by
Huber et al.~\cite{Huber2008MFI}. As required by the construction above, the
refined upper bound never exceeds the single-Gaussian bound. The gap between
these bounds varies as the fifth component moves between overlapping and
well-separated configurations.

\begin{figure}[t]
    \centering
    \includegraphics[width=\linewidth]{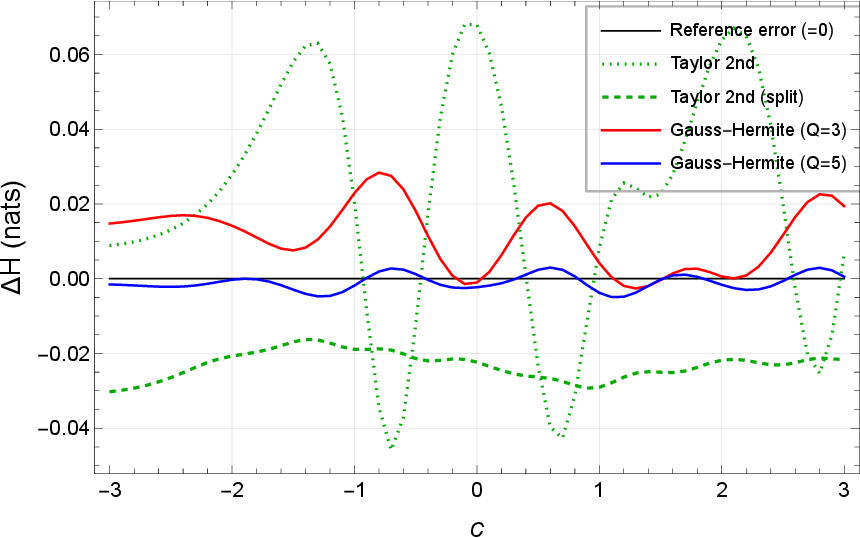}
\caption{
Signed entropy-estimation error relative to the independently
integrated numerical reference for the unsplit and split second-order Taylor approximations, GH3, and GH5 methods. The analytic lower and upper bounds
are omitted because they are intended to bracket the entropy rather
than approximate it symmetrically.
}
    \label{fig:multivariateDelta}
\end{figure}

Figure~\ref{fig:multivariateDelta} shows the signed
entropy-estimation errors defined in
\eqref{eq:multivariate_entropy_error}. Compared with GH3, GH5
exhibits a substantially smaller error amplitude over the parameter
sweep and remains close to the numerical reference at the scale of
the plot. The unsplit second-order Taylor approximation exhibits pronounced
oscillatory errors, whereas the split second-order Taylor approximation shows a
predominantly negative bias. The Gauss--Hermite estimator requires neither derivatives of
$\log g_c(x)$ nor component splitting.

\section{Hermite Surrogate for Repeated Action Evaluation}
\label{sec:optional_action_surrogate}

For repeated evaluation of an entropy-based objective over a continuous
action variable, we construct a Hermite polynomial surrogate in action
coordinates. This construction is separate from the state-space entropy
evaluator.

\subsection{Expected Posterior-Entropy Objective}
\label{subsec:expected_posterior_entropy}

Let $k$ denote the current decision epoch, let $X_k\in\mathbb{R}^{d_x}$
denote the state to be estimated, and let $\mathcal I_k^-$ denote all
information available before the next measurement is obtained. The
predicted belief density is defined as
\begin{equation}
b_k^-(x)
\triangleq
p(x\mid\mathcal I_k^-).
\label{eq:predicted_belief}
\end{equation}
Let $u\in\mathcal U_k\subset\mathbb{R}^{d_u}$ denote an admissible sensing
action selected before observing the prospective measurement
$Z_k\in\mathbb{R}^{d_z}$. The measurement model is specified by the conditional
density
\begin{equation}
p(z\mid x,u),
\qquad
z\in\mathbb{R}^{d_z}.
\label{eq:action_dependent_likelihood}
\end{equation}
The following conditional-independence assumption is imposed:
\begin{equation}
Z_k
\perp
\mathcal I_k^-
\mid
(X_k,u).
\label{eq:measurement_conditional_independence}
\end{equation}

We additionally assume that the sensing action does not alter the
predicted state density before the measurement is obtained:
\begin{equation}
p(x\mid\mathcal I_k^-,u)
=
b_k^-(x).
\label{eq:sensing_action_prior_assumption}
\end{equation}
This assumption is appropriate when $u$ controls only the sensing
configuration. If the action also affects the state transition, the
action-conditioned predicted density $b_k^-(x\mid u)$ must be used
instead.

By the conditional chain rule, the joint density of $X_k$ and $Z_k$,
conditioned on the available information and the selected action, is
\begin{align}
p(x,z\mid\mathcal I_k^-,u)
&=
p(z\mid x,\mathcal I_k^-,u)
p(x\mid\mathcal I_k^-,u)
\nonumber\\
&=
p(z\mid x,u)b_k^-(x),
\label{eq:predictive_joint_density}
\end{align}
where the second equality follows from
\eqref{eq:measurement_conditional_independence} and
\eqref{eq:sensing_action_prior_assumption}. Marginalizing the unknown
state gives the predictive measurement density
\begin{equation}
p_k(z\mid u)
\triangleq
p(z\mid\mathcal I_k^-,u)
=
\int_{\mathbb{R}^{d_x}}
p(z\mid x,u)b_k^-(x)\,dx.
\label{eq:predictive_measurement_density}
\end{equation}

For a hypothetical measurement realization $z$, Bayes' rule gives the
corresponding posterior belief
\begin{equation}
b_k^+(x\mid z,u)
\triangleq
p(x\mid z,\mathcal I_k^-,u)
=
\frac{
p(z\mid x,u)b_k^-(x)
}{
p_k(z\mid u)
},
\label{eq:hypothetical_posterior_belief}
\end{equation}
provided that $p_k(z\mid u)>0$. Here, $b_k^+(\cdot\mid z,u)$ denotes the hypothetical posterior
associated with candidate measurement $z$.

For the entropy evaluator developed in this paper to be applicable,
we assume that the posterior belief used for planning is represented
by a finite Gaussian mixture for every relevant pair $(z,u)$:
\begin{equation}
b_k^+(x\mid z,u)
=
\sum_{\ell=1}^{L_k^+(z,u)}
\omega_{\ell,k}^+(z,u)
\mathcal{N}
\left(
x;
\mu_{\ell,k}^+(z,u),
C_{\ell,k}^+(z,u)
\right),
\label{eq:posterior_gmm_representation}
\end{equation}
where $L_k^+(z,u)$ is the number of posterior components,
\begin{equation}
\omega_{\ell,k}^+(z,u)>0,
\qquad
\sum_{\ell=1}^{L_k^+(z,u)}
\omega_{\ell,k}^+(z,u)
=
1,
\end{equation}
and every posterior covariance matrix
$C_{\ell,k}^+(z,u)$ is positive definite. The entropy
$H(b_k^+(\cdot\mid z,u))$ can then be evaluated using the
Gauss--Hermite entropy estimator in
\eqref{eq:entropy_gh_estimator}.

Because the measurement is unknown when the action is selected, the
posterior entropy is averaged with respect to the predictive
measurement density in \eqref{eq:predictive_measurement_density}. The
expected posterior-entropy objective is therefore
\begin{align}
J_k(u)
&\triangleq
\mathbb{E}
\left[
H\!\left(
b_k^+(\cdot\mid Z_k,u)
\right)
\middle|
\mathcal I_k^-,u
\right]
\nonumber\\
&=
\int_{\mathbb{R}^{d_z}}
H\!\left(
b_k^+(\cdot\mid z,u)
\right)
p_k(z\mid u)\,dz.
\label{eq:expected_posterior_entropy}
\end{align}
An entropy-minimizing sensing action is then selected according to
\begin{equation}
u_k^\star
\in
\operatorname*{arg\,min}_{u\in\mathcal U_k}
J_k(u).
\label{eq:entropy_minimizing_action}
\end{equation}
After an actual measurement $z_k$ has been observed, the posterior is
$b_k^+(\cdot\mid z_k,u_k)$, and no outer expectation over $Z_k$ is
required.

\subsection{Action-Dependent Objective and Local Coordinates}

Let $u_{k,c}\in\mathcal U_k$ denote a nominal action at decision epoch
$k$. We introduce a smooth and injective coordinate map
\begin{equation}
\Psi_k:\mathbb{R}^{d_u}\rightarrow\mathcal U_k,
\qquad
u=\Psi_k(\xi),
\qquad
\Psi_k(0)=u_{k,c},
\label{eq:generic_action_mapping}
\end{equation}
where $\xi\in\mathbb{R}^{d_u}$ is a dimensionless coordinate vector and $d_u$
is the dimension of the action space. The admissibility condition
\begin{equation}
\Psi_k(\mathbb{R}^{d_u})
\subseteq
\mathcal U_k
\label{eq:generic_action_mapping_admissible}
\end{equation}
ensures that every coordinate value used in the Hermite projection is
mapped to an admissible physical action. Injectivity ensures that every
action in the image of $\Psi_k$ has a unique coordinate
representation.

If all action variables are unconstrained, the coordinate map may be
chosen as
\begin{equation}
\Psi_k(\xi)
=
u_{k,c}+S_{u,k}\xi,
\label{eq:generic_affine_action_mapping}
\end{equation}
where
\begin{equation}
S_{u,k}
\in
\mathbb{R}^{d_u\times d_u}
\label{eq:action_scaling_matrix_dimension}
\end{equation}
is a nonsingular action-scaling matrix. The matrix $S_{u,k}$ converts
the dimensionless coordinate vector $\xi$ into physical action
increments. It determines both the scale and, when it is nondiagonal,
the orientation of the action-space neighborhood represented by the
surrogate.

A common affine choice is
\begin{equation}
S_{u,k}
=
\operatorname{diag}
\left(
s_{u,k,1}^{\mathrm{phys}},
\ldots,
s_{u,k,d_u}^{\mathrm{phys}}
\right),
\qquad
s_{u,k,j}^{\mathrm{phys}}>0,
\label{eq:diagonal_action_scaling_matrix}
\end{equation}
where $s_{u,k,j}^{\mathrm{phys}}$ has the physical unit of the $j$th
action variable and specifies its characteristic local variation.

A nonsingular affine map has an unbounded image and therefore cannot
satisfy \eqref{eq:generic_action_mapping_admissible} for a bounded
action variable. Suppose that the $j$th action component satisfies
\begin{equation}
\underline u_{k,j}
<
u_j
<
\overline u_{k,j},
\label{eq:bounded_action_interval}
\end{equation}
and that its nominal value satisfies
\begin{equation}
\underline u_{k,j}
<
u_{k,c,j}
<
\overline u_{k,j}.
\label{eq:bounded_nominal_action}
\end{equation}
Define
\begin{equation}
\eta_{k,j}
\triangleq
\log
\left(
\frac{
u_{k,c,j}-\underline u_{k,j}
}{
\overline u_{k,j}-u_{k,c,j}
}
\right).
\label{eq:bounded_action_center_parameter}
\end{equation}
A smooth globally admissible coordinate map is
\begin{equation}
\Psi_{k,j}(\xi_j)
=
\underline u_{k,j}
+
\frac{
\overline u_{k,j}-\underline u_{k,j}
}{
1+
\exp
\left[
-\left(
\eta_{k,j}+\rho_{k,j}\xi_j
\right)
\right]
},
\qquad
\rho_{k,j}>0,
\label{eq:bounded_action_mapping}
\end{equation}
where $\rho_{k,j}$ is a dimensionless coordinate-scale parameter. This
map satisfies
\begin{equation}
\Psi_{k,j}(0)
=
u_{k,c,j}
\end{equation}
and
\begin{equation}
\underline u_{k,j}
<
\Psi_{k,j}(\xi_j)
<
\overline u_{k,j}
\qquad
\text{for every }
\xi_j\in\mathbb{R}.
\end{equation}

If a desired physical local scale
$s_{u,k,j}^{\mathrm{phys}}$ is specified by
\begin{equation}
\left.
\frac{d\Psi_{k,j}(\xi_j)}
{d\xi_j}
\right|_{\xi_j=0}
=
s_{u,k,j}^{\mathrm{phys}},
\label{eq:bounded_action_physical_scale_definition}
\end{equation}
then the corresponding dimensionless parameter is
\begin{equation}
\rho_{k,j}
=
\frac{
s_{u,k,j}^{\mathrm{phys}}
\left(
\overline u_{k,j}-\underline u_{k,j}
\right)
}{
\left(
u_{k,c,j}-\underline u_{k,j}
\right)
\left(
\overline u_{k,j}-u_{k,c,j}
\right)
}.
\label{eq:bounded_action_coordinate_scale}
\end{equation}
Affine maps for unconstrained variables and bounded maps of the form
\eqref{eq:bounded_action_mapping} may be combined when the action
vector contains both types of variables.

The objective in standardized action coordinates is defined as
\begin{equation}
F_k(\xi)
\triangleq
J_k\!\left(\Psi_k(\xi)\right),
\qquad
\xi\in\mathbb{R}^{d_u}.
\label{eq:standardized_action_objective}
\end{equation}
Thus, the coordinate map separates physical action units and
admissibility constraints from the dimensionless coordinates used in
the Hermite expansion.

The local surrogate is optimized over a prescribed coordinate domain
\begin{equation}
\mathcal D_k
\subset
\mathbb{R}^{d_u}.
\label{eq:local_coordinate_domain}
\end{equation}
The corresponding physical action neighborhood is
\begin{equation}
\mathcal U_k^{\mathrm{loc}}
\triangleq
\Psi_k(\mathcal D_k)
\subseteq
\mathcal U_k.
\label{eq:local_action_domain}
\end{equation}
The global coordinate map determines the admissible actions used for
the Hermite projection, whereas $\mathcal D_k$ restricts the region
over which the resulting polynomial surrogate is used for local
optimization.

\subsection{Hermite Projection in Action Coordinates}

The following construction uses a nonintrusive Hermite
polynomial-chaos, or pseudospectral, projection of the
action-dependent objective with respect to a standard normal
reference measure \cite{Xiu2010}. The standard normal measure is
introduced solely to define the Hermite projection in the
dimensionless action coordinates; it does not represent a
probability distribution of the physical sensing action. The
resulting polynomial is used as a local surrogate for repeated
objective evaluation and optimization.

Assume that
$F_k\in L^2(\mathcal{N}(0,I_{d_u}))$. This is an explicit regularity condition
on the action-dependent objective and the coordinate map. A sufficient
condition is that $J_k$ be bounded on the image of $\Psi_k$. Under this
condition, the Hermite coefficients of $F_k$ are
\begin{equation}
d_{k,\alpha}
=
\frac{1}{\alpha!}
\mathbb{E}_{\Xi\sim\mathcal{N}(0,I_{d_u})}
\left[
F_k(\Xi)\mathrm{He}_{\alpha}(\Xi)
\right],
\qquad
\alpha\in\mathbb N_0^{d_u}.
\label{eq:generic_action_hermite_coefficients}
\end{equation}
For sufficiently smooth $F_k$ whose required derivatives are integrable under
the standard normal measure, Gaussian integration by parts gives a derivative
interpretation of the low-order exact Hermite coefficients. In particular,
\begin{align}
d_{k,e_j}
&=\mathbb{E}\!\left[\frac{\partial F_k(\Xi)}{\partial \xi_j}\right],\\
d_{k,2e_j}
&=\frac{1}{2}\mathbb{E}\!\left[\frac{\partial^2 F_k(\Xi)}{\partial \xi_j^2}\right],\\
d_{k,e_j+e_\ell}
&=\mathbb{E}\!\left[\frac{\partial^2 F_k(\Xi)}{\partial \xi_j\partial\xi_\ell}\right],
\qquad j\neq\ell.
\end{align}
These identities expose the approximation principle of the Hermite surrogate.
A local Taylor quadratic is determined by the gradient and Hessian at the
nominal action, whereas the exact Hermite projection is determined by
Gaussian-weighted information over the standardized action space. The
implemented surrogate evaluates these projection coefficients directly from
objective values at Gauss--Hermite nodes. Consequently, the spatially
distributed objective evaluations are part of the Hermite projection itself,
rather than numerical-differentiation samples.

A total-order-$R$ surrogate is
\begin{equation}
\widetilde F_{k,R}(\xi)
=
\sum_{|\alpha|\leq R}
d_{k,\alpha}\mathrm{He}_{\alpha}(\xi).
\label{eq:generic_action_hermite_surrogate_exact}
\end{equation}
The number of retained basis functions is
\begin{equation}
N_R
=
\binom{d_u+R}{R}.
\label{eq:generic_action_basis_count}
\end{equation}
Unlike the entropy calculation, where Gauss--Hermite quadrature is applied
directly to a Gaussian expectation, the action-space surrogate explicitly
retains Hermite terms up to order $R$ to approximate $F_k(\xi)$ over the
local action domain.

Let
$\{(T_m^{(u)},W_m^{(u)})\}_{m=1}^{M_u}$
be a $d_u$-dimensional Gauss--Hermite rule. The coefficients in
\eqref{eq:generic_action_hermite_coefficients} are approximated by
\begin{equation}
\widehat d_{k,\alpha}
=
\frac{1}{\alpha!\,\pi^{d_u/2}}
\sum_{m=1}^{M_u}
W_m^{(u)}
F_k\!\left(\sqrt{2}\,T_m^{(u)}\right)
\mathrm{He}_{\alpha}\!\left(\sqrt{2}\,T_m^{(u)}\right).
\label{eq:generic_action_hermite_coefficients_gh}
\end{equation}
Because \eqref{eq:generic_action_hermite_coefficients_gh} replaces the exact
Gaussian projection integral by a finite quadrature rule,
$\widehat d_{k,\alpha}$ need not equal the exact projection coefficient
$d_{k,\alpha}$. Unresolved higher-order variation in $F_k$ can therefore enter
the computed low-order coefficients through quadrature error (aliasing). The
polynomial truncation error and the quadrature error are distinct. In the
radar benchmark, $R=2$ and $Q_u=3$ are fixed and the resulting surrogate
accuracy is assessed directly on independent validation actions.

Equivalently, direct objective evaluations are performed at the admissible
actions
\begin{equation}
u_{k,m}
=
\Psi_k\!\left(\sqrt{2}\,T_m^{(u)}\right),
\qquad
m=1,\ldots,M_u.
\label{eq:generic_action_quadrature_actions}
\end{equation}
For a full tensor-product rule with $Q_u$ nodes per action dimension,
\begin{equation}
M_u
=
Q_u^{d_u}.
\label{eq:generic_action_tensor_node_count}
\end{equation}
The surrogate is specified by the action-coordinate map $\Psi_k$, polynomial
order $R$, and quadrature order $Q_u$. The map $\Psi_k$ and quadrature order
$Q_u$ determine the extent and locations of the neighborhood samples, while
$R$ determines the retained polynomial space. The projection coefficients are
formed directly from objective values rather than from numerical derivative
estimates.
The computable surrogate is therefore
\begin{equation}
\widehat F_{k,R}(\xi)
=
\sum_{|\alpha|\leq R}
\widehat d_{k,\alpha}\mathrm{He}_{\alpha}(\xi).
\label{eq:generic_action_hermite_surrogate}
\end{equation}

For an entropy-minimization objective, the surrogate action is selected as
\begin{equation}
\widehat\xi_k
\in
\operatorname*{arg\,min}_{\xi\in\mathcal D_k}
\widehat F_{k,R}(\xi),
\qquad
\widehat u_k
=
\Psi_k(\widehat\xi_k).
\label{eq:generic_action_surrogate_selection}
\end{equation}
If the policy uses negative entropy or an information reward, the
equivalent maximization form is used. For a finite candidate-action set
\begin{equation}
\left\{
u_{k,n}
\right\}_{n=1}^{N_u}
\subset
\mathcal U_k^{\mathrm{loc}},
\end{equation}
the injectivity of $\Psi_k$ gives the corresponding coordinates
\begin{equation}
\xi_{k,n}
=
\Psi_k^{-1}(u_{k,n}),
\qquad
n=1,\ldots,N_u.
\end{equation}
The candidate actions can then be ranked by evaluating
$\widehat F_{k,R}(\xi_{k,n})$.

\subsection{Computational Cost}

Let $C_J$ denote the cost of one direct evaluation of $J_k(u)$. This cost may
include belief propagation, construction of an action-dependent posterior, an
outer expectation over possible measurements, and one or more state-space
Gaussian-mixture entropy evaluations. If direct optimization requires
$N_{\mathrm{dir}}$ objective evaluations, its cost is approximately
\begin{equation}
C_{\mathrm{direct}}
\approx
N_{\mathrm{dir}}C_J.
\label{eq:generic_direct_action_cost}
\end{equation}

The surrogate requires $M_u$ direct objective evaluations, projection of those
values onto $N_R$ basis functions, and repeated polynomial evaluations during
surrogate optimization. If $N_{\mathrm{surr}}$ denotes the number of surrogate
objective evaluations used by the optimizer, the total cost can be written as
\begin{equation}
C_{\mathrm{surr}}
\approx
M_uC_J
+
O\!\left(
M_uN_R
+
N_{\mathrm{surr}}N_R
\right).
\label{eq:generic_action_surrogate_cost}
\end{equation}
The construction can be beneficial when
$M_u\ll N_{\mathrm{dir}}$, the direct objective cost $C_J$ is large relative
to polynomial evaluation, and the objective is sufficiently smooth over the
selected action neighborhood.

The action-space surrogate reduces only the number of repeated evaluations of
$J_k$. It does not reduce the cost of an individual state-space entropy
calculation or of any outer expectation included in $J_k$. Those calculations
remain part of each of the $M_u$ direct evaluations used to construct the
surrogate.

\subsection{Applicability and Limitations}

The surrogate is local to the current objective, nominal action, coordinate
map, and action neighborhood. It should therefore be reconstructed when the
underlying belief or model changes appreciably. Its accuracy depends on the
smoothness of $J_k\circ\Psi_k$ over the region sampled by the action-space
quadrature rule and over the optimization domain $\mathcal D_k$.

Low-order polynomial surrogates may be inaccurate in the presence of hard
field-of-view boundaries, discontinuous gating, discrete association changes,
component pruning or merging, mode switching, or other nonsmooth operations.
A practical implementation should assess the surrogate at independent
validation actions and revert to direct evaluation or rebuild the surrogate
when the validation error exceeds a prescribed tolerance.

The tensor-product cost $M_u=Q_u^{d_u}$ grows exponentially with the action
dimension. Sparse-grid or adaptive rules may therefore be required when
$d_u$ is not small. In addition, increasing the polynomial order $R$ is useful
only when the available quadrature rule and direct objective evaluations can
resolve the corresponding higher-order variation.

\section{Closed-Loop Radar Pointing Benchmark for the Action Surrogate}
\label{sec:radar_action_benchmark}

The action-space surrogate is evaluated in a closed-loop radar pointing
problem with fixed dwell time. The comparison isolates beam-pointing
decisions while all methods use the same sensing dwell time. The purpose of
this benchmark is to evaluate the action-space surrogate, not to compare
alternative state-space entropy estimators; the same Gauss--Hermite
state-space entropy evaluator is held fixed across all action-selection
methods.
\subsection{Target, Radar, and Gaussian-Mixture Tracker}
\label{subsec:radar_scenario}

The tracked target has Cartesian state
\begin{equation}
x
=
\begin{bmatrix}
p_x & p_y & p_z & v_x & v_y & v_z
\end{bmatrix}^{\top}
\in\mathbb{R}^6.
\label{eq:radar_state}
\end{equation}
The radar measures range, azimuth, and elevation,
\begin{equation}
z
=
h(x)+\nu,
\qquad
\nu\sim\mathcal{N}(0,R(x,u)),
\label{eq:radar_measurement}
\end{equation}
where the pointing action is
\begin{equation}
u
=
\begin{bmatrix}
\phi_b & \theta_b
\end{bmatrix}^{\top},
\label{eq:radar_pointing_action}
\end{equation}
with beam azimuth $\phi_b$ and beam elevation $\theta_b$. The dwell
time is fixed at $\tau=1\,\mathrm{s}$ throughout the reported
benchmark.

Let $r$, $\phi$, and $\theta$ denote the target range, azimuth, and
elevation, and define the pointing errors
\begin{equation}
\Delta\phi=\phi-\phi_b,
\qquad
\Delta\theta=\theta-\theta_b.
\label{eq:radar_pointing_errors}
\end{equation}
The measurement covariance is modeled as
\begin{equation}
R(x,u)
=
\left(\frac{r}{r_0}\right)^{\kappa}
\exp\left[
\frac{\Delta\phi^2}{2\sigma_{\phi,b}^2}
+
\frac{\Delta\theta^2}{2\sigma_{\theta,b}^2}
\right]
\frac{\tau_0}{\tau}
R_0,
\label{eq:radar_measurement_covariance}
\end{equation}
where
\begin{equation}
R_0
=
\operatorname{diag}
\left(
(35\,\mathrm{m})^2,
\left(0.45\frac{\pi}{180}\right)^2,
\left(0.45\frac{\pi}{180}\right)^2
\right),
\end{equation}
where the angular covariance entries are expressed in radians squared; the
equivalent one-standard-deviation angular scale is $0.45^\circ$.
Furthermore, $r_0=6\,\mathrm{km}$, $\kappa=2$,
$\sigma_{\phi,b}=\sigma_{\theta,b}=3^\circ$, and
$\tau_0=1\,\mathrm{s}$. Thus, measurement uncertainty grows with
range and off-boresight angle and decreases with dwell time.

A smooth detection-probability model is used to avoid a discontinuity
in the planning objective. Define
\begin{equation}
\gamma^2(x,u)
=
\left(\frac{\Delta\phi}{\phi_g}\right)^2
+
\left(\frac{\Delta\theta}{\theta_g}\right)^2,
\qquad
\phi_g=\theta_g=2^\circ,
\label{eq:radar_gate_metric}
\end{equation}
and let
\begin{equation}
\sigma_{\mathrm{L}}(a)
=
\frac{1}{1+\exp(-a)}.
\end{equation}
The detection probability is
\begin{equation}
p_D(x,u)
=
p_{D,\max}
\sigma_{\mathrm{L}}\!\left(
k_a(1-\gamma^2(x,u))
\right)
\sigma_{\mathrm{L}}\!\left(
k_r\left(1-\frac{r}{r_{\mathrm{det}}}\right)
\right),
\label{eq:radar_detection_probability}
\end{equation}
with $p_{D,\max}=0.995$, $k_a=7$, $k_r=10$, and
$r_{\mathrm{det}}=25\,\mathrm{km}$.

Because the radar measurement model is nonlinear, each Gaussian component is
updated approximately by an extended-Kalman linearization of $h(x)$ about the
component mean. The component means and covariances are updated with the
resulting Kalman gain, and the mixture weights are updated from the
corresponding Gaussian innovation likelihoods. Consequently, the posterior
GMM used by the planning and entropy calculations is an approximate filtering
representation rather than the exact posterior of the nonlinear measurement
model.

The Gaussian-mixture tracker uses three maneuver hypotheses: left turn,
straight motion, and right turn. Their initial weights are
$0.25$, $0.50$, and $0.25$, respectively, and the fixed turn-rate
hypotheses are $+6^\circ/\mathrm{s}$, $0$, and
$-6^\circ/\mathrm{s}$. The truth maneuver is allowed to exceed these
fixed model rates, producing controlled model mismatch as the maneuver
severity increases.

The initial target slant range is $6\,\mathrm{km}$, the elevation is
$15^\circ$, and the velocity is approximately perpendicular to the
initial line of sight, with a nominal speed of
$265\,\mathrm{m/s}$. This crossing geometry produces appreciable
line-of-sight angular motion even before the evasive turn begins.
Table~\ref{tab:radar_benchmark_settings} summarizes the main settings.

\begin{table}[t]
\centering
\caption{Main settings of the closed-loop radar pointing benchmark.}
\label{tab:radar_benchmark_settings}
\begin{tabular}{ll}
\toprule
Quantity & Value \\
\midrule
Simulation step / duration & $0.1\,\mathrm{s}$ / $20\,\mathrm{s}$ \\
Replanning period & $1.0\,\mathrm{s}$ \\
Common center-pointing burn-in & $0\le t<3\,\mathrm{s}$ \\
Method-specific policy active & $t\ge 3\,\mathrm{s}$ \\
Evasive maneuver start & $t=5\,\mathrm{s}$ \\
Maneuver severity & $2,4,6,8,10,12^\circ/\mathrm{s}$ \\
Initial slant range / elevation & $6\,\mathrm{km}$ / $15^\circ$ \\
Initial crossing angle / speed & $90^\circ$ / $265\,\mathrm{m/s}$ \\
Fixed dwell time & $1.0\,\mathrm{s}$ \\
Azimuth / elevation gate & $2^\circ$ / $2^\circ$ \\
Action-map asymptotic bounds & $\pm5^\circ$ / $\pm3^\circ$ \\
Standardized optimization domain & $[-1.5,1.5]^2$ \\
State-space quadrature order & $Q_x=2$ in $d_x=6$ \\
Action-space quadrature order & $Q_u=3$ in $d_u=2$ \\
Taylor derivative step & $h_T^\star=1.953125\times10^{-4}$ (offline calibrated) \\
Taylor calibration snapshots & $72$ snapshots from $12$ pilot trials \\
Monte Carlo trials per severity & $50$ \\
\bottomrule
\end{tabular}
\end{table}

\subsection{Planning Objective and Compared Policies}
\label{subsec:radar_planning_objective}

At each replanning epoch, the nominal beam center
$u_{k,c}=[\phi_{k,c},\theta_{k,c}]^\top$ is obtained from the line of
sight to the current Gaussian-mixture mean. A smooth centered map from
standardized coordinates to physical beam angles is used:
\begin{align}
\phi_b
&=
\phi_{k,c}
+
\phi_{\lim}
\tanh\left(
\frac{s_\phi}{\phi_{\lim}}\xi_1
\right),
\nonumber\\
\theta_b
&=
\theta_{k,c}
+
\theta_{\lim}
\tanh\left(
\frac{s_\theta}{\theta_{\lim}}\xi_2
\right),
\label{eq:radar_action_map}
\end{align}
where
\begin{equation}
\phi_{\lim}=5^\circ,
\qquad
\theta_{\lim}=3^\circ,
\qquad
s_\phi=1.5^\circ,
\qquad
s_\theta=1.0^\circ.
\end{equation}
The quantities $s_\phi$ and $s_\theta$ define the common action-coordinate
map used by all compared action-selection methods. They set the physical scale
associated with a standardized coordinate and therefore determine the physical
sampling geometry of both surrogate constructions.
The surrogate is optimized over
$\mathcal D=[-1.5,1.5]^2$. Under \eqref{eq:radar_action_map}, this
corresponds to maximum searched offsets of approximately
$\pm2.11^\circ$ in azimuth and $\pm1.39^\circ$ in elevation about the
nominal beam center.

The selected beam is held until the next replanning epoch. To avoid a
terminal-only planning bias, the planning objective is averaged over three
midpoint samples at offsets
\begin{equation}
\left\{\frac{1}{6},\frac{1}{2},\frac{5}{6}\right\}\,\mathrm{s}
\end{equation}
from the replanning epoch within the one-second action-hold interval; hence,
$N_t=3$.

For planning only, each predicted mixture component supplies a
noise-free hypothetical measurement at its predicted mean. At planning
sample $q$, let $\mu_{\ell,q}^-$ and $\omega_{\ell,q}^-$ denote the
predicted mean and weight of component $\ell$, and define
\begin{equation}
z_{\ell,q}^{\mathrm{PIMS}}
\triangleq
h\!\left(\mu_{\ell,q}^-\right).
\label{eq:radar_pims_measurement}
\end{equation}
The collection of these hypothetical measurements is referred to as
the \emph{predicted ideal measurement set} (PIMS). PIMS is used only
inside the planning objective; the closed-loop tracker receives
stochastic detections and noisy measurements.

Let $u(\xi)$ denote the beam-pointing action defined by
\eqref{eq:radar_action_map}, and define
\begin{equation}
p_{D,\ell,q}(\xi)
\triangleq
p_D\!\left(
\mu_{\ell,q}^-,
u(\xi)
\right).
\label{eq:radar_pims_detection_probability}
\end{equation}
Let
$b_{\ell,q}^{+,\mathrm{PIMS}}(\cdot\mid\xi)$ denote the posterior
belief obtained from $b_q^-$ using the hypothetical measurement
$z_{\ell,q}^{\mathrm{PIMS}}$ under action $u(\xi)$.

The generic objective in \eqref{eq:standardized_action_objective} involves an
expectation over prospective measurements. In the radar benchmark this
expectation is replaced by the PIMS construction below and augmented by a
missed-detection penalty. We denote the resulting directly evaluable,
benchmark-specific action objective by $F_k^{\mathrm{PIMS}}(\xi)$. The action
surrogates in the remainder of this section approximate this quantity, not the
exact outer expectation in \eqref{eq:expected_posterior_entropy}.

The implemented PIMS planning objective is
\begin{align}
F_k^{\mathrm{PIMS}}(\xi)
\triangleq
\frac{1}{N_t}
\sum_{q=1}^{N_t}
\sum_{\ell}
\omega_{\ell,q}^-
\Big[
&
p_{D,\ell,q}(\xi)
H\!\left(
b_{\ell,q}^{+,\mathrm{PIMS}}(\cdot\mid\xi)
\right)
\nonumber\\
&+
\left(
1-p_{D,\ell,q}(\xi)
\right)
\left(
H(b_q^-)+\lambda_{\mathrm{miss}}
\right)
\Big],
\label{eq:radar_pims_objective}
\end{align}
where $\lambda_{\mathrm{miss}}=3$ is a fixed planning penalty used to
make a missed detection unfavorable relative to the entropy variations in the
benchmark; it is a design weight rather than a physical radar parameter. The state-space GMM entropies in
\eqref{eq:radar_pims_objective} are evaluated by the deterministic
Gauss--Hermite estimator developed above. The reported benchmark uses
$Q_x=2$, giving $2^6=64$ state-space quadrature nodes per component
expectation. The same state-space entropy evaluator is used for all action-selection
methods.

The primary action-surrogate comparison uses a second-order local Taylor
surrogate and a second-order Hermite surrogate for the PIMS objective
$F_k^{\mathrm{PIMS}}$. A $9\times9$ direct search,
requiring 81 objective evaluations, is retained as a computational
reference. The two quadratic surrogates differ in the information used to
construct their coefficients. The Taylor surrogate estimates the local
quadratic jet of $F_k^{\mathrm{PIMS}}$ at $\xi=0$ from central finite
differences, whereas the Hermite surrogate estimates Gaussian-weighted
projection coefficients from objective values distributed over the
standardized action neighborhood. The Taylor finite-difference step is
calibrated offline on an independent set of center-pointing pilot states using
derivative-stability criteria.

Let $g_h$ and $B_h$ denote the finite-difference gradient and Hessian computed
with step $h$. For each adjacent pair $h$ and $h/2$, the calibration records
\begin{align}
E_g(h)
&=
\frac{\lVert g_h-g_{h/2}\rVert_2}
{\max\!\left(1,\lVert g_{h/2}\rVert_2\right)},
\label{eq:radar_taylor_gradient_stability}\\
E_B(h)
&=
\frac{\lVert B_h-B_{h/2}\rVert_F}
{\max\!\left(1,\lVert B_{h/2}\rVert_F\right)}.
\label{eq:radar_taylor_hessian_stability}
\end{align}
The calibration uses a halving sequence from $0.4$ down to
$6.1035\times10^{-6}$ on 72 planning snapshots from 12 independent
pilot trials: three pilot trials at each of $2$, $6$, $10$, and
$12^\circ/\mathrm{s}$, with snapshots taken at
$t=3,5,8,12,15,$ and $18\,\mathrm{s}$. The calibration pilot trials use random realizations independent of those used in the closed-loop evaluation. The selected step is the finer member of the adjacent
pair minimizing the 90th percentile of $E_B(h)$, with the corresponding
gradient criterion used as a tie breaker. The minimum occurs for
$3.90625\times10^{-4}\rightarrow1.953125\times10^{-4}$, for which the
90th-percentile Hessian and gradient changes are $2.07\times10^{-6}$ and
$1.01\times10^{-6}$, respectively. The final Taylor baseline therefore uses
\begin{equation}
h_T^\star
=
1.953125\times10^{-4}.
\label{eq:radar_taylor_step}
\end{equation}
The calibration is performed offline and is excluded from the reported
online action-selection time. Once $h_T^\star$ is fixed, the
two-dimensional Taylor model requires nine direct objective evaluations per
replanning step.

The second-order Hermite surrogate uses the total-order-two Hermite
construction in Section~\ref{sec:optional_action_surrogate} to
$F_k^{\mathrm{PIMS}}$. A tensor-product $Q_u=3$
Gauss--Hermite rule in two dimensions likewise requires
\begin{equation}
M_u=Q_u^{d_u}=3^2=9
\end{equation}
direct objective evaluations. For a standard normal expectation, the
one-dimensional standardized nodes are
\begin{equation}
\{-\sqrt{3},\,0,\,+\sqrt{3}\}.
\label{eq:radar_action_gh_nodes}
\end{equation}
For this $d_u=2$, $Q_u=3$ rule, the normalized one-dimensional
standard-normal quadrature weights are
\begin{equation}
\left\{\frac{1}{6},\,\frac{2}{3},\,\frac{1}{6}\right\}.
\label{eq:radar_action_gh_normalized_weights}
\end{equation}
Under the resulting nine-point tensor-product discrete measure, the tensor
Hermite modes satisfying $0\leq\alpha_j\leq2$ are mutually orthogonal.
Consequently, retaining the six modes with total degree $|\alpha|\leq2$ is
equivalent to the discrete Gaussian-weighted least-squares projection of the
nine directly evaluated objective values onto the total-degree-two polynomial
space. The quadratic Hermite surrogate need not interpolate all nine values;
the statement is a weighted-projection equivalence.
The physical sampling radius of these nodes is determined by the coordinate
scales in \eqref{eq:radar_action_map} and the bounded $\tanh$ map. The outer quadrature nodes
$\xi_j=\pm\sqrt{3}$ extend beyond the optimization interval
$[-1.5,1.5]$ and supply the neighborhood values used in the Hermite
projection. The surrogate is optimized over $\mathcal D$, and the bounded map
\eqref{eq:radar_action_map} maps every quadrature node to a physically
admissible action. The resulting Hermite quadratic is therefore constructed from nine
Gauss--Hermite evaluations of the objective. These evaluations approximate
the Gaussian-weighted projection integrals defining the Hermite coefficients.
By contrast, the nine evaluations used by the Taylor implementation serve to
estimate the gradient and Hessian at $\xi=0$ by finite differences. Thus,
although both implementations require nine direct objective evaluations per
replanning step, the evaluations have different mathematical roles. Their
resulting quadratic surrogates are optimized over the same standardized
candidate domain.

A companion state-space quadrature sensitivity calculation was performed at
the most severe maneuver, $12^\circ/\mathrm{s}$. For the second-order Hermite
surrogate, increasing $Q_x$ from 2 to 3, corresponding to an increase from
$2^6=64$ to $3^6=729$ nodes per component expectation, changed the surrogate
RMSE only from 0.556854 to 0.556855 and the mean optimizer regret from 0.010233
to 0.010233. The detection rate, inside-gate rate, temporary-loss rate, and
terminal-loss rate were unchanged to the reported precision, while the mean
online time in this separate sensitivity run increased from approximately
$38.8$ to $75.0\,\mathrm{ms}$.
Thus, the reported Hermite-surrogate policy metrics at this maneuver severity
are insensitive to the tested increase in $Q_x$. The calculation is a
policy-metric sensitivity study with respect to the state-space quadrature
order.

A separate diagnostic reference search measures optimizer regret and
standardized action error. It combines a denser direct grid with bounded
multistart refinement and is evaluated every second replanning epoch. Its cost
is excluded from the reported online action-selection time. The online
$9\times9$ direct grid serves as a computational reference, while the
diagnostic search provides the finer action-space reference used in the regret
metrics; the mean regret of the online direct grid relative to this reference
is approximately $1.8\times10^{-3}$ over the six reported severities.

\subsection{Monte Carlo Design and Paired Analysis}
\label{subsec:radar_paired_analysis}

All policies use the same center-pointing controller for the first
$3\,\mathrm{s}$, so that the comparison starts from a common tracked
condition. Method-specific action selection begins at
$t=3\,\mathrm{s}$ and the evasive maneuver begins at
$t=5\,\mathrm{s}$. Let $s_{\deg}$ denote the reported maneuver severity in
degrees per second and define
\begin{equation}
s=\frac{\pi}{180}s_{\deg}
\end{equation}
in radians per second. With $\tau=t-5\,\mathrm{s}$, the commanded yaw rate is
\begin{equation}
\dot\psi_{\mathrm{cmd}}(\tau)
=
\begin{cases}
0, & \tau<0,\\
s, & 0\le\tau<3.5\,\mathrm{s},\\
-1.15s, & 3.5\le\tau<7.0\,\mathrm{s},\\
0.70s, & 7.0\le\tau<10.5\,\mathrm{s},\\
0, & \tau\ge10.5\,\mathrm{s}.
\end{cases}
\label{eq:radar_maneuver_schedule}
\end{equation}
The reported severity $s_{\deg}$ is varied over
$2,4,6,8,10,$ and $12^\circ/\mathrm{s}$.

For the primary surrogate comparison, 50 Monte Carlo trials are performed
for each surrogate at each severity, giving
\begin{equation}
6\times50\times2=600
\end{equation}
closed-loop runs. A paired common-random-number design is used: within each
trial, the Taylor and Hermite methods share the same realization of the truth
process, initial filter perturbation, detection events, and measurement noise.
Separate random-number streams are used for these stochastic quantities. The
finite-difference calibration is completed beforehand using an independent
set of random realizations.

Surrogate diagnostics are evaluated at every second replanning epoch. At each
such epoch, a fresh set of 32 validation actions is sampled uniformly over
$\mathcal D=[-1.5,1.5]^2$ using a random-number stream separate from the
closed-loop process and measurement streams. Surrogate RMSE is computed from
the difference between the quadratic surrogate and the directly evaluated
PIMS objective on these actions. Spearman rank correlation is
computed from the same validation evaluations. Let
\begin{equation}
\Delta F_k^{\mathrm{val}}
\triangleq
\max_j F_k^{\mathrm{PIMS}}(\xi_{k,j}^{\mathrm{val}})-\min_j F_k^{\mathrm{PIMS}}(\xi_{k,j}^{\mathrm{val}})
\end{equation}
denote the objective range over the validation set. A range-normalized RMSE is defined only for validation epochs satisfying
$\Delta F_k^{\mathrm{val}}\geq0.25$:
\begin{equation}
\mathrm{NRMSE}_k
\triangleq
\frac{\mathrm{RMSE}_k}{\Delta F_k^{\mathrm{val}}}.
\label{eq:radar_surrogate_nrmse}
\end{equation}
The threshold $0.25$ is used solely as a numerical guard against unstable
normalization on nearly flat validation surfaces; it is not tuned using
closed-loop performance. Validation epochs with
$\Delta F_k^{\mathrm{val}}<0.25$ are retained for the absolute RMSE and
rank-correlation diagnostics but are excluded from the NRMSE average. Let
$\widehat{\xi}_k$ denote the action selected by the surrogate and
$\xi_k^{\mathrm{ref}}$ the action returned by the diagnostic reference search.
The optimizer regret and standardized action error are defined as
\begin{align}
r_k
&\triangleq
\max\!\left\{
0,\,
F_k^{\mathrm{PIMS}}(\widehat{\xi}_k)
-
F_k^{\mathrm{PIMS}}(\xi_k^{\mathrm{ref}})
\right\},
\label{eq:radar_optimizer_regret}
\\
e_{\xi,k}
&\triangleq
\left\|
\widehat{\xi}_k-\xi_k^{\mathrm{ref}}
\right\|_2.
\label{eq:radar_standardized_action_error}
\end{align}
The normalized beam error is
\begin{equation}
e_{b,k}
\triangleq
\sqrt{\gamma^2(x_k,u_k)},
\label{eq:radar_normalized_beam_error}
\end{equation}
where $\gamma^2(x,u)$ is defined in \eqref{eq:radar_gate_metric}; a sample is
inside the pointing gate when $e_{b,k}\leq1$. Detection rate, inside-gate
rate, mean normalized beam error, and position RMSE are computed over the
method-specific policy interval $t\geq3\,\mathrm{s}$, while online
action-selection time is recorded at each method-specific replanning epoch.

A temporary loss event is declared after the common burn-in if either five
consecutive detections are missed or the position error reaches
$750\,\mathrm{m}$. Terminal loss is assessed from the final miss streak and
the median position error over the last $2\,\mathrm{s}$. Reacquisition
requires three consecutive detections with position error below the loss
threshold.

Because the same stochastic realization is reused across methods,
statistical inference is performed on paired differences. For a metric
for which smaller values are better, the reported paired improvement is
\begin{equation}
\Delta_m
=
m_{\mathrm{Taylor}}
-
m_{\mathrm{Hermite}},
\label{eq:radar_paired_improvement_small}
\end{equation}
whereas for a metric for which larger values are better,
\begin{equation}
\Delta_m
=
m_{\mathrm{Hermite}}
-
m_{\mathrm{Taylor}}.
\label{eq:radar_paired_improvement_large}
\end{equation}
Hence, positive improvement always favors the Hermite surrogate. 
The primary overall effect estimate is the mean trial-cluster paired
difference, with uncertainty reported using 20,000 cluster-paired bootstrap
resamples. The bootstrap confidence interval is the primary inferential
summary because it targets the reported mean effect directly. Wilcoxon signed-rank tests are reported as
secondary distribution-robust checks for continuous cluster differences;
severity-specific signed-rank tests use Holm correction. Binary loss outcomes
are compared severity by severity using the exact McNemar test. This separation
is important because a confidence interval for the mean paired effect and a
rank-based test need not lead to identical threshold decisions for a skewed or
discrete cluster-difference distribution.

The six severity levels are not treated as 300 independent paired
samples in the overall comparison. Instead, each trial is first
averaged across severity and the resulting 50 trial clusters are used
as the independent statistical units. This clustered construction
preserves the common-random-number design across maneuver severity.

\subsection{Results}
\label{subsec:radar_results}

Table~\ref{tab:paired_action_surrogate} gives the primary trial-cluster paired
comparison between the derivative-calibrated Taylor surrogate and the Hermite
surrogate. The mean surrogate RMSE decreases from 2.7190 to 0.5610, while the
mean range-normalized surrogate RMSE decreases from 0.8612 to 0.1788. The
paired NRMSE reduction is 0.6824 with 95\% bootstrap confidence interval
$[0.6610,0.7043]$. All 50 trial-cluster differences favor the Hermite
surrogate for RMSE, NRMSE, optimizer regret, and standardized action error.
The mean validation-objective ranges are 3.1413 for the derivative-calibrated
Taylor runs and 3.1429 for the Hermite runs, so the large difference in
absolute RMSE is not explained by a different validation-objective scale.
The fractions of diagnostic epochs excluded from NRMSE aggregation by the
$\Delta F_k^{\mathrm{val}}<0.25$ guard are only 1.44\% and 1.15\%,
respectively. The mean Spearman rank correlation between the surrogate and
directly evaluated objective increases from 0.6554 to 0.9464.

\begin{table}[t]
\centering
\footnotesize
\setlength{\tabcolsep}{2.5pt}
\caption{Primary overall paired Monte Carlo comparison of the
derivative-calibrated second-order Taylor surrogate and the second-order
Hermite surrogate. Positive improvement favors the Hermite surrogate.
Confidence intervals are 20,000-resample trial-cluster paired-bootstrap
95\% intervals.}
\label{tab:paired_action_surrogate}
\begin{tabular}{lccc}
\toprule
Metric & Taylor surrogate & Hermite surrogate & Improvement [95\% CI] \\
\midrule
Surrogate RMSE
& 2.7190 & 0.5610 & $2.1580\,[2.0871,2.2330]$ \\
Normalized surrogate RMSE
& 0.8612 & 0.1788 & $0.6824\,[0.6610,0.7043]$ \\
Optimizer regret
& 0.1321 & 0.0095 & $0.1225\,[0.0989,0.1509]$ \\
Standardized action error
& 0.6027 & 0.1838 & $0.4189\,[0.3978,0.4439]$ \\
Detection rate
& 0.8850 & 0.9429 & $0.0579\,[0.0510,0.0650]$ \\
Temporary loss-event rate
& 0.2267 & 0.1667 & $0.0600\,[-0.0067,0.1333]$ \\
\bottomrule
\end{tabular}
\end{table}

\begin{figure}[t]
\centering
\begin{minipage}{0.49\linewidth}
\centering
\includegraphics[width=\linewidth]{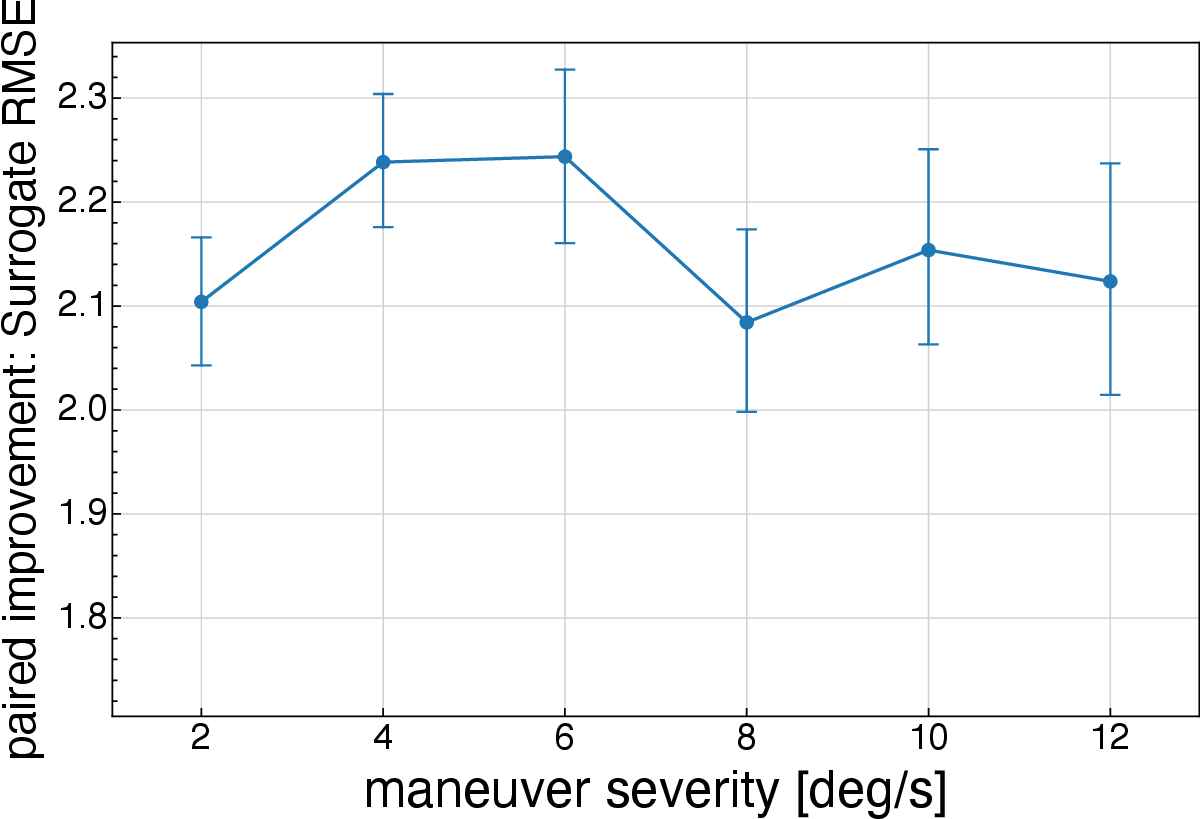}\\[-1mm]
{\small (a) Surrogate RMSE}
\end{minipage}\hfill
\begin{minipage}{0.49\linewidth}
\centering
\includegraphics[width=\linewidth]{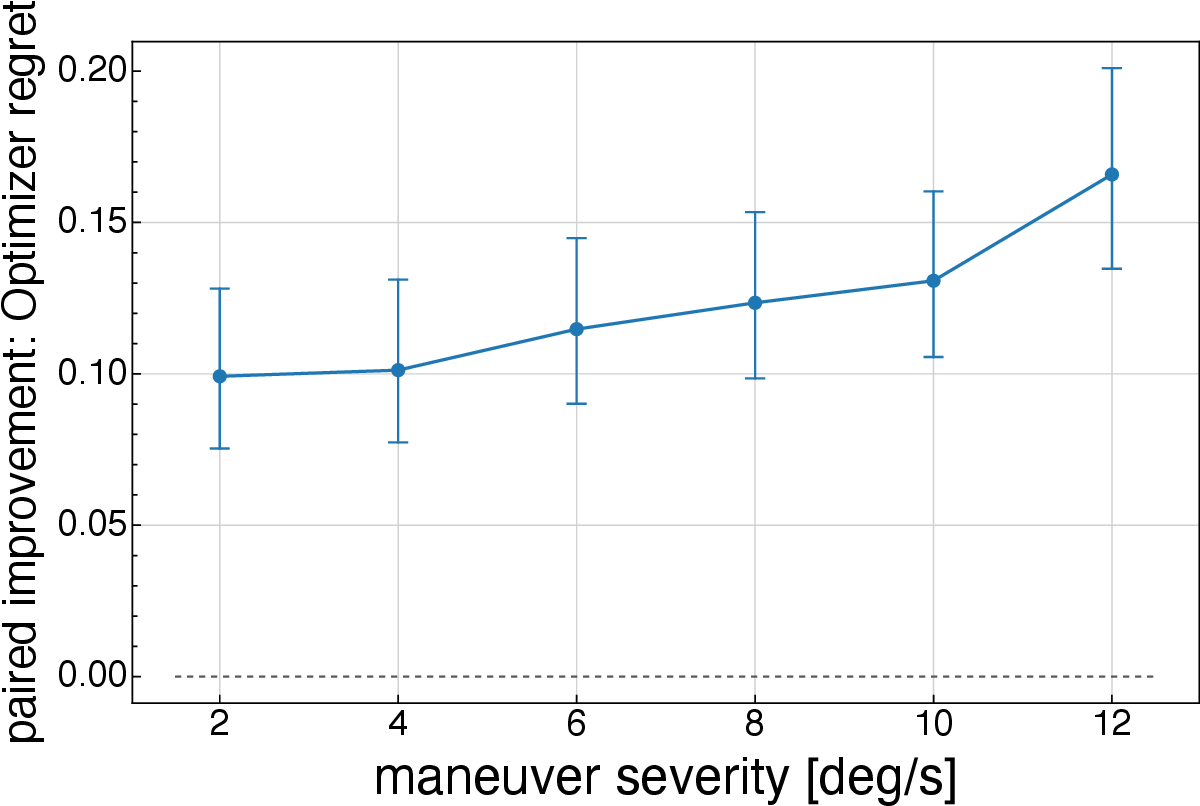}\\[-1mm]
{\small (b) Optimizer regret}
\end{minipage}
\caption{Paired improvement of the second-order Hermite surrogate relative
to the derivative-calibrated second-order Taylor surrogate. Positive values
favor the Hermite surrogate. Error bars are paired-bootstrap 95\% confidence
intervals.}
\label{fig:radar_paired_numerical}
\end{figure}

Figure~\ref{fig:radar_paired_numerical} shows that the improvement in
surrogate RMSE and optimizer regret remains positive over all six maneuver
severities. The RMSE reduction stays near 2.1--2.2 across the full sweep,
whereas the regret reduction grows from approximately 0.099 at
$2^\circ/\mathrm{s}$ to 0.166 at $12^\circ/\mathrm{s}$. This distinction is
useful: surface-approximation error and the quality of the selected minimizer
are related but are not identical diagnostics.

For the primary derivative-calibrated Taylor comparison, the numerical
improvement carries through to the closed-loop sensing performance. The overall
detection rate increases from 0.8850 for the Taylor surrogate to 0.9429 for the
Hermite surrogate, a paired increase of 0.0579 with 95\% bootstrap confidence
interval $[0.0510,0.0650]$. The corresponding overall inside-gate rate
increases from 0.9072 to 0.9534, while the mean normalized beam error decreases
from 0.6026 to 0.4687. The detection-rate advantage increases with maneuver severity, from
approximately 0.037 at $2^\circ/\mathrm{s}$ to 0.091 at
$12^\circ/\mathrm{s}$.

Temporary loss events decrease from 0.2267 for the derivative-calibrated
Taylor surrogate to 0.1667 for the Hermite surrogate. The overall paired mean
absolute risk reduction is 0.0600, but its primary cluster-bootstrap interval
$[-0.0067,0.1333]$ includes zero, so no overall loss-prevention claim is made. At $12^\circ/\mathrm{s}$ the observed
temporary-loss rate decreases from 0.44 to 0.24. The severity-specific exact
McNemar test is $p=0.021$ before multiplicity correction and $p=0.128$ after
Holm correction across the six severities. Terminal loss remains rare
(0.033 for the Taylor surrogate and 0.020 for the Hermite surrogate overall), and the corresponding
paired interval includes zero. The overall mean position RMSE decreases from
$81.1$ to $76.5\,\mathrm{m}$, but this secondary effect is much less uniform
across trials than the surrogate, action, and pointing metrics.
\begin{figure}[t]
\centering
\begin{minipage}{0.49\linewidth}
\centering
\includegraphics[width=\linewidth]{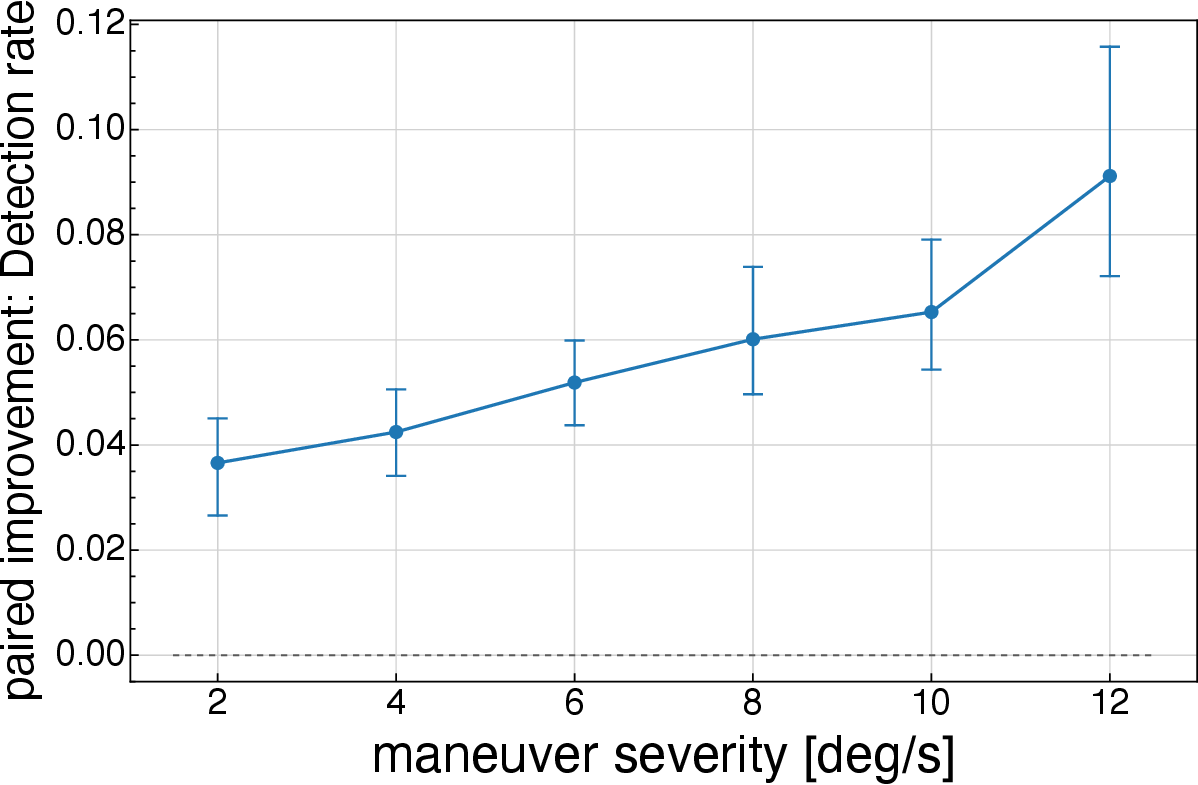}\\[-1mm]
{\small (a) Detection rate}
\end{minipage}\hfill
\begin{minipage}{0.49\linewidth}
\centering
\includegraphics[width=\linewidth]{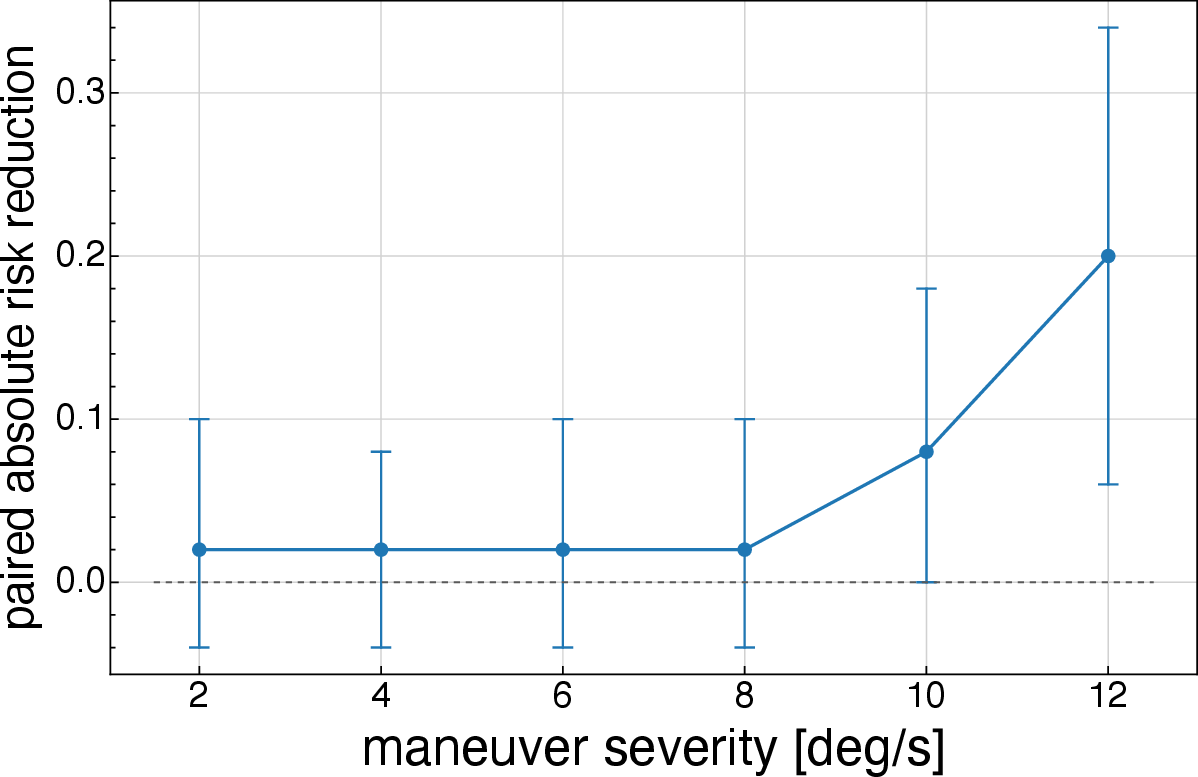}\\[-1mm]
{\small (b) Temporary loss-event rate}
\end{minipage}
\caption{Closed-loop paired effects of the second-order Hermite surrogate
relative to the derivative-calibrated second-order Taylor surrogate. For detection rate, the plotted
quantity is
$P_{D,\mathrm{Hermite}}-P_{D,\mathrm{Taylor}}$.
For temporary loss, the plotted quantity is the absolute risk reduction
$P_{\mathrm{loss,Taylor}}-P_{\mathrm{loss,Hermite}}$.
Positive values favor the Hermite surrogate.}
\label{fig:radar_paired_closed_loop}
\end{figure}

The computational comparison is summarized in
Table~\ref{tab:radar_online_cost}. The calibrated second-order Taylor and
Hermite surrogates each use nine direct objective evaluations per replanning
step, at their respective sampling locations, and have essentially identical
measured online times, approximately $39.4\,\mathrm{ms}$ in the paired rerun.
The 81-point direct search requires about
$359\,\mathrm{ms}$ in the corresponding direct-search timing run. Thus, the dominant computational
saving is the reduction from 81 direct objective evaluations to 9; changing
the Taylor finite-difference step cannot reduce that call count. Absolute
timings are implementation- and hardware-dependent, so the evaluation count is
the more transferable measure.

\begin{table}[t]
\centering
\small
\setlength{\tabcolsep}{3pt}
\caption{Online action-selection cost. Diagnostic-reference and offline
Taylor-step calibration computations are excluded.}
\label{tab:radar_online_cost}
\begin{tabular}{@{}lccc@{}}
\toprule
Method & Calls/replan & Mean time [ms] &
\shortstack{Approx.\ speedup\\vs.\ direct} \\
\midrule
Direct grid
& 81 & 358.9 & 1.0 \\[6pt]

\shortstack[l]{2nd-order Taylor\\surrogate ($h_T^\star$)}
& 9 & 39.4 & 9.1 \\[6pt]

\shortstack[l]{2nd-order Hermite\\surrogate}
& 9 & 39.4 & 9.1 \\
\bottomrule
\end{tabular}
\end{table}

Relative to the derivative-calibrated local Taylor baseline, the Hermite
surrogate yields substantially lower surrogate error and optimizer regret and
better pointing metrics at the same nine-call online objective cost. The two
methods differ in the information used to construct their quadratic models:
the Taylor surrogate estimates local derivatives at the nominal action,
whereas the Hermite surrogate evaluates a Gaussian-weighted projection from
objective values distributed over the standardized action neighborhood. The
closed-loop results show that the latter construction provides a substantially
more accurate and effective action surrogate for the tested radar problem.

\section{Discussion}
\label{sec:discussion}

The entropy calculation consists of standardizing each Gaussian component,
followed by direct Gauss--Hermite quadrature of the resulting Gaussian
expectations. The Hermite basis provides an equivalent interpretation: each
expectation corresponds to the constant Hermite coefficient of the
standardized integrand, but no truncated Hermite expansion is required for
the entropy calculation. The quadrature order therefore serves as the
numerical resolution parameter of the entropy evaluator.

The entropy benchmarks show that low-order Gauss--Hermite rules closely track
the numerical references on the tested one- and two-dimensional mixtures. Quadrature accuracy
depends on the standardized mixture geometry, including component separation,
covariance anisotropy, and overlap. Successive-order differences such as
$|\widehat H_Q-\widehat H_{Q+2}|$ provide a practical numerical diagnostic of
quadrature resolution on a given mixture family.

The numerical comparisons reported here use the Huber Taylor constructions,
analytic bounds, and independent numerical references. A complementary
comparison with Monte Carlo entropy estimation can be made at matched
computational cost. For an $L$-component Gaussian mixture with $M$
Gauss--Hermite nodes used for each Gaussian expectation, such a comparison is
naturally based on approximately $LM$ evaluations of $\log g(x)$ or on
measured runtime rather than on the quadrature-node count alone.

The action-space construction serves a different purpose from the
entropy-only quadrature. It retains higher-order Hermite terms to approximate
the action-dependent objective and is naturally interpreted as a nonintrusive
Hermite polynomial-chaos or pseudospectral surrogate. The closed-loop radar
benchmark evaluates the quadratic case. Its central distinction from the
local Taylor model is information locality. The Taylor coefficients represent
derivatives at the nominal action, whereas the Hermite coefficients represent
Gaussian-weighted neighborhood information. The Stein identities in
Section~\ref{sec:optional_action_surrogate} express the exact first- and
second-order Hermite coefficients as Gaussian averages of derivatives, while
the implemented projection coefficients are computed directly from objective
values at fixed Gauss--Hermite nodes.
The spatially distributed Gauss--Hermite evaluations are therefore part of
the Hermite approximation principle. Their role is to approximate projection
integrals, not to approximate pointwise derivatives. This distinction is
central to the action-space comparison: the Taylor and Hermite surrogates use
the same number of direct objective evaluations, while extracting different
information from those evaluations.

The $Q_x=2$ versus $Q_x=3$ companion calculation at
$12^\circ/\mathrm{s}$ provides a sensitivity check on the state-space
quadrature order. For the Hermite surrogate, the surrogate RMSE, optimizer regret,
detection rate, inside-gate rate, and loss outcomes are unchanged to the
reported precision when the six-dimensional rule is
increased from $2^6$ to $3^6$ nodes per component expectation, while
the measured online time in this separate sensitivity run increases from about
$38.8$ to $75.0\,\mathrm{ms}$. For this tested case, increasing the state-space quadrature order from
$Q_x=2$ to $Q_x=3$ does not materially change the reported performance
metrics, while approximately doubling the measured online time.

The radar benchmark is a steerable-beam sensor-management model in which
pointing controls the detection probability and measurement covariance. Within
this model, the primary paired comparison shows lower surrogate error, lower
optimizer regret, smaller standardized action error, higher detection and
inside-gate rates, and lower normalized beam error for the Hermite surrogate
relative to the derivative-calibrated local Taylor baseline. Temporary and terminal loss outcomes are less statistically
resolved, and the position-RMSE effect is secondary and less uniform across
trials.

The main computational limitation of full tensor-product Gauss--Hermite rules
is exponential growth with dimension. Strongly anisotropic or difficult
mixtures may also require additional nodes. Sparse grids, dimension-adaptive
rules, and problem-specific transformations are natural extensions. The same decomposition into Gaussian expectations can be used for
cross-entropy and Kullback--Leibler divergence calculations when the second
density can be evaluated at the quadrature nodes; these extensions preserve the distinction
between exact decomposition and numerical quadrature approximation.

\section{Conclusion}
\label{sec:conclusion}

We evaluated Gaussian-mixture differential entropy using direct
componentwise Gauss--Hermite quadrature. Standardizing each Gaussian
component reduces the entropy calculation to Gaussian expectations,
which are evaluated directly by quadrature. The resulting evaluator
requires neither derivative evaluation nor component splitting. On the
one- and two-dimensional benchmark families considered here, low-order
Gauss--Hermite rules agree closely with numerical integration references.

For repeated optimization over continuous actions, we constructed a
Hermite polynomial surrogate in action space. In a closed-loop radar sensor-management benchmark,
its second-order form achieves substantially lower surrogate error and
optimizer regret than a second-order Taylor surrogate based on local
derivatives at the nominal action, while both methods use nine direct
objective evaluations per replanning step. The Hermite surrogate also improves detection, inside-gate, and
beam-pointing metrics in the tested benchmark.

The entropy accuracy results are specific to the one- and two-dimensional
benchmark families considered here and do not establish a general convergence
rate with quadrature order.

Future work includes sparse-grid and adaptive quadrature for higher dimensions,
systematic convergence studies across different mixture geometries, and
higher-dimensional action surrogates under constrained and nonsmooth
objectives.

\section*{Acknowledgments}
This work was partially supported by the KIST Institutional Program.
\bibliographystyle{plain}
\bibliography{references_merged}

@article{BostromRostEtAl2021,
  author  = {Per Bostr{\"o}m-Rost and Daniel Axehill and Gustaf Hendeby},
  title   = {Sensor Management for Search and Track Using the {Poisson
             Multi-Bernoulli} Mixture Filter},
  journal = {IEEE Transactions on Aerospace and Electronic Systems},
  volume  = {57},
  number  = {5},
  pages   = {2771--2783},
  year    = {2021},
  doi     = {10.1109/TAES.2021.3061802}
}

@article{HernandezEtAl2024,
  author  = {Marcel L. Hernandez and {\'A}ngel F. Garc{\'i}a-Fern{\'a}ndez
             and Simon Maskell},
  title   = {Nonmyopic Sensor Control for Target Search and Track Using
             a Sample-Based {GOSPA} Implementation},
  journal = {IEEE Transactions on Aerospace and Electronic Systems},
  volume  = {60},
  number  = {1},
  pages   = {387--404},
  year    = {2024},
  doi     = {10.1109/TAES.2023.3324908}
}

@article{JonesEtAl2024,
  author  = {George Jones and {\'A}ngel F. Garc{\'i}a-Fern{\'a}ndez
             and Christian Blackman},
  title   = {Nonmyopic {GOSPA}-Driven {G}aussian {B}ernoulli Sensor Management},
  journal = {IEEE Transactions on Aerospace and Electronic Systems},
  volume  = {60},
  number  = {6},
  pages   = {7628--7642},
  year    = {2024},
  doi     = {10.1109/TAES.2024.3418750}
}

@article{Parzen1962,
  author  = {Emanuel Parzen},
  title   = {On Estimation of a Probability Density Function and Mode},
  journal = {The Annals of Mathematical Statistics},
  volume  = {33},
  number  = {3},
  pages   = {1065--1076},
  year    = {1962},
  doi     = {10.1214/aoms/1177704472}
}

@book{Silverman1986,
  author    = {Bernard W. Silverman},
  title     = {Density Estimation for Statistics and Data Analysis},
  publisher = {Chapman and Hall},
  address   = {London},
  year      = {1986}
}

@article{JoudehSkoric2026,
  author  = {Basheer Joudeh and Boris {\v{S}}kori{\'c}},
  title   = {Numerical Evaluation of {G}aussian Mixture Entropy},
  journal = {Entropy},
  year    = {2026},
  volume  = {28},
  number  = {4},
  pages   = {381},
  doi     = {10.3390/e28040381}
}

@inproceedings{SaalHeessVijayakumar2011,
  author    = {Hannes P. Saal and Nicolas M. O. Heess and Sethu Vijayakumar},
  title     = {Multimodal Nonlinear Filtering Using {G}auss--{H}ermite Quadrature},
  booktitle = {Machine Learning and Knowledge Discovery in Databases},
  editor    = {Dimitrios Gunopulos and Thomas Hofmann and Donato Malerba
               and Michalis Vazirgiannis},
  pages     = {81--96},
  publisher = {Springer},
  year      = {2011},
  doi       = {10.1007/978-3-642-23808-6_6}
}

@misc{FuruyaEtAl2022EntropyError,
  author       = {Takashi Furuya and Hiroyuki Kusumoto and Koichi Taniguchi and Naoya Kanno and Kazuma Suetake},
  title        = {Theoretical Error Analysis of Entropy Approximation for {G}aussian Mixtures},
  year         = {2022},
  eprint       = {2202.13059},
  archivePrefix= {arXiv},
  primaryClass = {stat.ML},
  note         = {arXiv:2202.13059}
}

@article{NielsenNock2017,
  author  = {Frank Nielsen and Richard Nock},
  title   = {MaxEnt Upper Bounds for the Differential Entropy of Univariate Continuous Distributions},
  journal = {IEEE Signal Processing Letters},
  year    = {2017},
  volume  = {24},
  number  = {4},
  pages   = {402--406},
  doi     = {10.1109/LSP.2017.2666792},
  note    = {arXiv:1612.02954}
}

@inproceedings{DahlkePacheco2023,
 author = {Dahlke, Caleb and Pacheco, Jason},
 booktitle = {Advances in Neural Information Processing Systems},
 doi = {10.52202/075280-3299},
 editor = {A. Oh and T. Naumann and A. Globerson and K. Saenko and M. Hardt and S. Levine},
 pages = {75469--75490},
 publisher = {Curran Associates, Inc.},
 title = {On Convergence of Polynomial Approximations to the {G}aussian Mixture Entropy},
 volume = {36},
 year = {2023}
}

@article{Shannon1948,
  author  = {Shannon, Claude E.},
  title   = {A Mathematical Theory of Communication},
  journal = {Bell System Technical Journal},
  year    = {1948},
  volume  = {27},
  number  = {3},
  pages   = {379--423},
  doi     = {10.1002/j.1538-7305.1948.tb01338.x},
  note    = {Part II appeared in vol. 27, no. 4, pp. 623--656 (1948).}
}

@inproceedings{Huber2008MFI,
  author    = {Huber, Marco F. and Bailey, Tim and Durrant-Whyte, Hugh and Hanebeck, Uwe D.},
  title     = {On Entropy Approximation for {G}aussian Mixture Random Vectors},
  booktitle = {Proceedings of the IEEE International Conference on Multisensor Fusion and Integration for Intelligent Systems (MFI)},
  year      = {2008},
  pages     = {181--188},
  doi       = {10.1109/MFI.2008.4648062}
}

@book{Gautschi2004,
  author    = {Gautschi, Walter},
  title     = {Orthogonal Polynomials: Computation and Approximation},
  publisher = {Oxford University Press},
  year      = {2004},
  doi       = {10.1093/oso/9780198506720.001.0001},
  isbn      = {9780198506720}
}

@book{Xiu2010,
  author    = {Xiu, Dongbin},
  title     = {Numerical Methods for Stochastic Computations: A Spectral Method Approach},
  publisher = {Princeton University Press},
  year      = {2010},
  isbn      = {9780691142128}
}

@book{AbramowitzStegun1965,
  editor    = {Abramowitz, Milton and Stegun, Irene A.},
  title     = {Handbook of Mathematical Functions: With Formulas, Graphs, and Mathematical Tables},
  publisher = {Dover Publications},
  year      = {1965},
  isbn      = {9780486612720},
  note      = {Reprint of National Bureau of Standards Applied Mathematics Series 55 (1964).}
}

@book{CoverThomas1991,
  author    = {Cover, Thomas M. and Thomas, Joy A.},
  title     = {Elements of Information Theory},
  publisher = {Wiley-Interscience},
  address   = {New York},
  year      = {1991},
  isbn      = {9780471062592}
}

@inproceedings{ManyikaDurrantWhyte1992,
  author    = {Manyika, James M. and Durrant-Whyte, Hugh F.},
  title     = {Information-theoretic approach to management in decentralized data fusion},
  booktitle = {Proc. SPIE 1828, Sensor Fusion V},
  year      = {1992},
  pages     = {202--213}
}

@article{ViolaWells1997,
  author  = {Viola, Paul and Wells III, William M.},
  title   = {Alignment by Maximization of Mutual Information},
  journal = {International Journal of Computer Vision},
  year    = {1997},
  volume  = {24},
  pages   = {137--154},
  doi     = {10.1023/A:1007958904918}
}

@inproceedings{HersheyOlsen2007,
  author    = {Hershey, John R. and Olsen, Peder A.},
  title     = {Approximating the {K}ullback-{L}eibler Divergence between {G}aussian Mixture Models},
  booktitle = {2007 IEEE International Conference on Acoustics, Speech and Signal Processing (ICASSP)},
  year      = {2007},
  volume    = {4},
  pages     = {IV-317--IV-320},
  doi       = {10.1109/ICASSP.2007.366913}
}

@inproceedings{GoldbergerGordonGreenspan2003,
  author    = {Goldberger, Jacob and Gordon, Shiri and Greenspan, Hayit},
  title     = {An Efficient Image Similarity Measure based on Approximations of {KL}-Divergence between Two {G}aussian Mixtures},
  booktitle = {Proceedings Ninth IEEE International Conference on Computer Vision (ICCV)},
  year      = {2003},
  volume    = {1},
  pages     = {487--493},
  doi       = {10.1109/ICCV.2003.1238387}
}

@article{Runnalls2007,
  author  = {Runnalls, Andrew R.},
  title   = {{Kullback-Leibler} Approach to {G}aussian Mixture Reduction},
  journal = {IEEE Transactions on Aerospace and Electronic Systems},
  year    = {2007},
  volume  = {43},
  number  = {3},
  pages   = {989--999},
  doi     = {10.1109/TAES.2007.4383588}
}

@inproceedings{HanebeckBriechleRauh2003,
    author       = {Hanebeck, U. D. and Briechle, K. and Rauh, A.},
    year         = {2003},
    title        = {Progressive Bayes: A New Framework for Nonlinear State Estimation},
    pages        = {256 - 267},
    booktitle    = {Multisensor, Multisource Information Fusion: Architectures, Algorithms, and Applications},
    doi          = {10.1117/12.487806},
    publisher    = {{Society of Photo-optical Instrumentation Engineers (SPIE)}},
    isbn         = {978-0-81-944959-7},
    series       = {Proceedings of Spie},
    language     = {english},
    volume       = {5099}
}

@article{KullbackLeibler1951,
  author  = {Kullback, Solomon and Leibler, Richard A.},
  title   = {On Information and Sufficiency},
  journal = {The Annals of Mathematical Statistics},
  year    = {1951},
  volume  = {22},
  number  = {1},
  pages   = {79--86},
  doi     = {10.1214/aoms/1177729694}
}

@inproceedings{Salmond1989IEEColloquium,
  author    = {Salmond, David J.},
  title     = {Mixture reduction algorithms for target tracking},
  booktitle = {IEE Colloquium on State Estimation in Aerospace and Tracking Applications},
  year      = {1989},
  address   = {London, UK},
  month     = dec,
  pages     = {7/1--7/4},
  publisher = {IEE}
}

@article{MelbourneEtAl2022,
  author  = {James Melbourne and Saurav Talukdar and Shreyas Bhaban and Mokshay Madiman and Murti V. Salapaka},
  title   = {The Differential Entropy of Mixtures: New Bounds and Applications},
  journal = {IEEE Transactions on Information Theory},
  volume  = {68},
  number  = {4},
  pages   = {2123--2146},
  year    = {2022},
  doi     = {10.1109/TIT.2022.3140661}
}

@article{Zobay2014,
  author  = {Oliver Zobay},
  title   = {Variational Bayesian Inference with {G}aussian-Mixture Approximations},
  journal = {Electronic Journal of Statistics},
  volume  = {8},
  number  = {1},
  pages   = {355--389},
  year    = {2014},
  doi     = {10.1214/14-EJS887}
}
\end{document}